\documentclass[11pt]{article}

\usepackage[margin=1in]{geometry}

\usepackage{graphicx}
\graphicspath{{../figures/}{figures/}}

\usepackage{amsmath,amssymb,amsfonts,amsthm}
\usepackage{listings}
\usepackage{caption}
\usepackage{float}
\usepackage{booktabs}
\usepackage{xcolor}
\usepackage[colorlinks=true,linkcolor=blue,citecolor=blue,urlcolor=blue]{hyperref}
\usepackage{algorithm}
\usepackage{algpseudocode}
\usepackage{multirow}
\usepackage{enumitem}
\usepackage[numbers,sort&compress]{natbib}
\usepackage{authblk}

\begin{document}

\title{\textbf{Interpretable epistemic uncertainty decomposition in sequential generative models via polynomial chaos surrogates}}

\author[2]{Ram\'on Nartallo-Kaluarachchi}
\author[3]{Shashanka Ubaru}
\author[1]{Ma{\l}gorzata~J. Zimon}
\author[4]{Dongsung Huh}
\author[1]{Robert Manson-Sawko}
\author[3]{Lior Horesh\thanks{Corresponding author: \texttt{lhoresh@us.ibm.com}}}
\author[5]{Yoshua Bengio}

\affil[1]{IBM Research Europe, Daresbury, United Kingdom}
\affil[2]{Mathematical Institute, University of Oxford, United Kingdom}
\affil[3]{IBM Research, Thomas J. Watson Research Center, USA}
\affil[4]{MIT--IBM Watson AI Lab, Massachusetts, USA}
\affil[5]{Universit\'e de Montr\'eal, Mila--Quebec AI Institute, Montreal, Quebec, Canada}

\date{}
\maketitle
\addtocontents{toc}{\protect\setcounter{tocdepth}{-5}}

\begin{abstract}
Sequential generative models conditioned on uncertain rewards are central to AI-driven scientific discovery, yet the epistemic uncertainty they inherit from imperfect reward estimates remains unquantified. We propagate this uncertainty through generative flow networks (GFlowNets) by fitting polynomial chaos expansions (PCEs) to small ensembles of trained models. The PCE coefficients yield analytical Sobol sensitivity indices, providing the first interpretable decomposition of \emph{which reward components drive which generative decisions}, a capability unavailable from deep ensembles, Bayesian neural networks, or Monte Carlo dropout. Convergence guarantees are established theoretically and four of five are formally verified in the Lean~4 proof assistant. Across three real-world tasks the framework reveals actionable structure invisible to ensembles alone. On the Doyle--Dreher Buchwald--Hartwig dataset~\cite{ahneman2018predicting} catalyst selection is robust ($D_{\mathrm{catalyst}}\approx 71$) while additive selection is fragile ($D_{\mathrm{additive}}\approx 179$, $2.5\times$ higher). In fragment-based molecular design the linker position is the most sensitive ($D_{\mathrm{linker}}\approx 28$) while decoration positions are the most robust ($D\approx 14$--$18$), reversing the conventional scaffold-robust / decoration-fragile assumption. On the Sachs protein signalling network, MAPK-cascade edges and PKA/PKC hub edges separate into distinct sensitivity regimes, providing a targeted map for perturbation experiments. Calibration coverage at the 95\% level reaches 0.97--1.00 across the dominant steps, and the surrogate evaluates 10{,}000 policy samples in milliseconds -- $10^{3}$--$10^{4}\times$ faster than exhaustive retraining.
\end{abstract}


\section{Introduction}\label{sec:intro}

Generative models that construct outputs through sequential decision-making, including generative flow networks (GFlowNets)~\cite{bengio2021flow,bengio2023gflownet}, autoregressive transformers~\cite{vaswani2017attention}, and diffusion models~\cite{sohl2015deep}, have become central tools across scientific discovery~\cite{jain2023gflownets}, combinatorial optimisation~\cite{zhang2023flows}, and biological design~\cite{jain2022biological}. These models are often described as uncertainty-aware because they define probability distributions over outputs. However, this aleatoric uncertainty, arising from the stochastic generation process, is distinct from epistemic uncertainty, which reflects ignorance about the model itself~\cite{gal2016uncertainty}.

Epistemic uncertainty is especially consequential when generative models are conditioned on reward or scoring functions that are themselves uncertain. In molecular design, reward functions are typically parameterised by neural network surrogates trained on limited experimental data~\cite{jain2023gflownets,chakraborti2025personalized}. In chemical reaction optimisation, yield predictors are trained on high-throughput but incomplete datasets~\cite{ahneman2018predicting}. In Bayesian structure learning, the marginal likelihood used to score candidate graphs depends on finite observational datasets~\cite{deleu2022bayesian}. In each case, the learned generative policy inherits uncertainty from the reward, yet standard GFlowNets offer no mechanism to express or quantify it.

Existing approaches to epistemic uncertainty in deep learning, including Bayesian neural networks~\cite{blundell2015weight,mackay1995probable} and deep ensembles~\cite{lakshminarayanan2017simple,gal2016dropout}, face well-known challenges. Bayesian methods are difficult to scale; ensemble methods demand repeated training at substantial computational cost. Crucially, none of these approaches provide \emph{an interpretable decomposition} of uncertainty: they can quantify that a prediction is uncertain, but not \emph{why} it is uncertain or \emph{which components} of the input uncertainty are responsible.

We draw inspiration from the field of uncertainty quantification (UQ) in computational engineering, where surrogate models approximate expensive simulators to enable efficient uncertainty propagation~\cite{sullivan2015introduction,conti2024multi}. A cornerstone technique is the polynomial chaos expansion (PCE)~\cite{wiener1938homogeneous,xiu2002wiener}, which represents a quantity of interest as a polynomial function of random inputs. Uniquely among surrogate methods, the PCE coefficients enable \emph{analytical} computation of Sobol sensitivity indices~\cite{sudret2008global}, providing an exact variance-based decomposition of which input uncertainties drive which outputs. Recent extensions to high-dimensional settings via deep PCE~\cite{exenberger2026deep}, PCE combined with variational autoencoders for representation learning~\cite{shustin2026pcenet}, and neural network architectures based on arbitrary polynomial chaos~\cite{oladyshkin2023deep} have expanded the applicability of these ideas.

Here we propose a surrogate modelling framework for quantifying and decomposing epistemic uncertainty in GFlowNets (Fig.~\ref{fig:framework}). Our approach proceeds in three stages: (i)~train a modest ensemble of GFlowNets under varied reward conditions, (ii)~learn a low-dimensional parameterisation of the reward function space, and (iii)~fit a PCE that maps from this low-dimensional representation to the policy distribution at each step of a trajectory. The fitted surrogate enables both inexpensive Monte Carlo estimation of policy uncertainty and, our central contribution, \emph{analytical Sobol sensitivity indices} that decompose policy uncertainty into contributions from each component of reward uncertainty.

We make four contributions:
\begin{enumerate}
    \item A general framework for propagating epistemic uncertainty through conditioned sequential generative models, with PCE surrogates fitted to small training ensembles.
    \item Formal convergence guarantees for the surrogate and its Sobol indices, with four of five key properties verified in Lean~4 (T1, T3a, T3b, T4); two formalisation gaps remain on ratio continuity and softmax Jacobian bounds.
    \item The first analytical sensitivity decomposition of policy uncertainty in generative models, revealing which reward components drive which generative decisions.
    \item Demonstrations on seven tasks of increasing complexity: three real-world applications -- chemical reaction condition optimisation on the Doyle--Dreher Buchwald--Hartwig dataset~\cite{ahneman2018predicting} (with Sobol indices directly addressing the Chuang--Keiser debate~\cite{chuang2018comment}), fragment-based molecular design with uncertain drug-likeness reward, and Bayesian causal discovery on the Sachs protein signalling network~\cite{sachs2005causal} -- together with four further validation tasks (discrete and continuous grid-worlds, symbolic regression, and a GRU-based LLM GFlowNet with uncertain process reward model).
\end{enumerate}

\section{Results}\label{sec:results}

\subsection{Surrogate modelling framework}\label{sec:framework}

We consider a GFlowNet trained on a reward function $R$ sampled from a distribution $\mathcal{R}$ that reflects epistemic uncertainty—each realisation of $R$ yields a different learned policy. Our goal is to characterise the marginal distribution over policies induced by $\mathcal{R}$, without exhaustive retraining.

\begin{figure}[t]
    \centering
    \includegraphics[width=\textwidth]{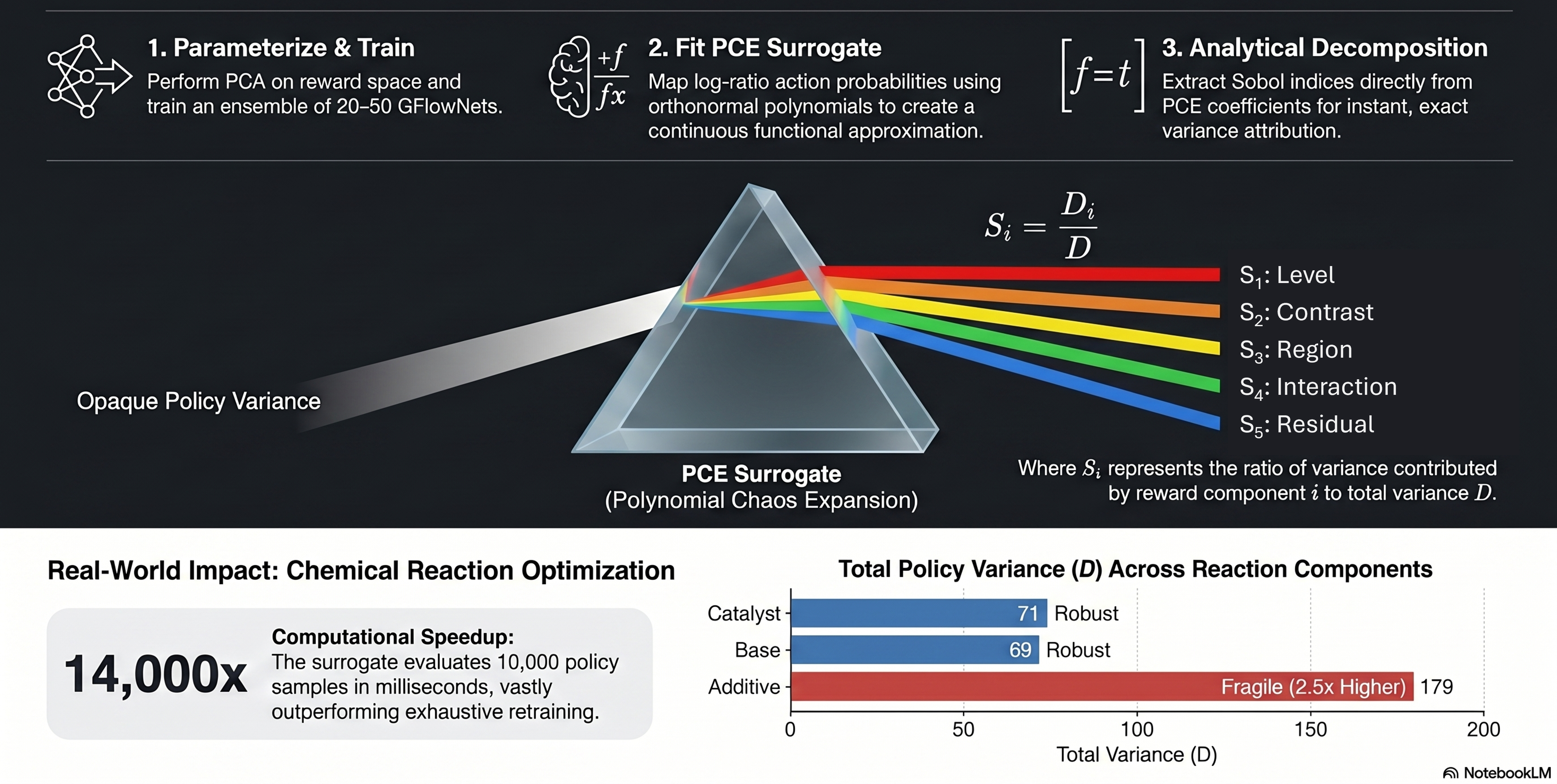}
    \caption{\textbf{The uncertainty prism: decoding generative decision-making.} The PCE surrogate acts as an interpretable prism that decomposes opaque policy variance into analytical Sobol sensitivity indices $S_i = D_i / D$, attributing uncertainty to individual reward principal components. Each outgoing ray is labelled by the canonical role of its principal component: $S_1$ \emph{level} (global offset of the reward surface), $S_2$ \emph{contrast} (peak-vs-bulk sharpness), $S_3$ \emph{region} (localised disagreement), $S_4$ \emph{interaction} (cross-factor coupling), $S_5$ \emph{residual} (fine-scale tail). These roles are intrinsic to the epistemic covariance of an ensemble reward surrogate and transfer unchanged across application domains. \textbf{Top:}~The three-stage pipeline: (1)~parameterise the reward space via PCA and train a small ensemble of GFlowNets, (2)~fit a PCE surrogate mapping log-ratio-transformed action probabilities through orthonormal polynomials, (3)~extract Sobol indices directly from the PCE coefficients. \textbf{Centre:}~The prism metaphor -- opaque policy variance enters, and interpretable Sobol indices exit, revealing which reward components drive which decisions. \textbf{Bottom:}~Real-world impact on the Buchwald--Hartwig reaction: total policy variance $D$ by reaction step reveals that catalyst ($D \approx 71$) and base ($D \approx 69$) selection are robust, while additive selection is fragile ($D \approx 179$, $2.5\times$ higher variance), a distinction invisible to ensemble methods alone. The surrogate evaluates 10{,}000 policy samples in milliseconds compared to exhaustive retraining (${\sim}14{,}000\times$ speedup).}
    \label{fig:framework}
\end{figure}

The framework has three components (Fig.~\ref{fig:framework}; see Methods for full details). First, a low-dimensional parameterisation $\boldsymbol{\mu} \in \mathbb{R}^d$ of the reward function is obtained via principal component analysis (PCA) or an analytical expansion (PCA is preferred because its orthogonal components yield independent Gaussian inputs, a prerequisite for valid analytical Sobol decomposition; see Methods). Second, a small ensemble of $L$ GFlowNets is trained, each on a distinct reward realisation, and the policy along a trajectory of interest is extracted. Third, for each action $k$ at trajectory step $t$, a PCE maps from $\boldsymbol{\mu}$ to the log-ratio-transformed action probability (see Methods, Eq.~\ref{eq:pce}), fitted by regularised regression. The surrogate then enables rapid Monte Carlo sampling and, uniquely, analytical computation of Sobol sensitivity indices from the PCE coefficients (Methods, Eq.~\ref{eq:sobol}).

\textbf{Why not simply analyse the raw ensemble?} If one trains $L$ GFlowNets, the ensemble already provides $L$ policy samples, and one might ask why a PCE surrogate is needed. Three reasons motivate the additional step. First, Sobol sensitivity indices require a smooth functional form: computing them from a finite-sample ensemble requires expensive Monte Carlo integration over the input space, which negates the computational advantage and introduces sampling noise, whereas the PCE yields exact analytical indices from its coefficients. Second, the PCE is a continuous functional approximation and can reveal structure, such as multimodality or non-monotone dependence, invisible in a finite sample. Third, the surrogate supports counterfactual queries, enabling evaluation of ``what if the uncertainty had a different structure?'' without retraining any GFlowNets.

\subsection{Reaction condition optimisation: Buchwald--Hartwig cross-coupling}\label{sec:buchwald}

A central question in machine learning for chemistry is whether learned models capture genuine chemical signals or statistical artifacts of the training set. Chuang and Keiser~\cite{chuang2018comment} demonstrated that the influential Doyle--Dreher yield predictor~\cite{ahneman2018predicting} for palladium-catalysed Buchwald--Hartwig amination contains strong substrate-identity effects that can confound interpretation of apparent model accuracy. The debate has remained largely unresolved because existing methods quantify \emph{aggregate} prediction uncertainty but cannot attribute it to specific reaction components. The Sobol sensitivity decomposition directly addresses this gap: a low Sobol index on a reaction component indicates that the GFlowNet recommendation is driven by a genuine chemical signal replicated across training subsets, while a high Sobol index flags a component whose recommendation changes with the training set, the hallmark of a statistical artifact.

We framed reaction condition selection as a 4-step GFlowNet construction: the agent sequentially selects the catalyst, base, aryl halide, and additive, with reward proportional to the predicted yield from a proxy model trained on a subset of the data. The Doyle--Dreher dataset~\cite{ahneman2018predicting} provides experimentally measured yields for 4,599 reactions spanning 16 aryl halides, 4 ligand/catalyst systems, 3 bases, and 24 isoxazole additives. By training the proxy on only 30\% of reactions, substantial epistemic uncertainty arises from unobserved combinations. We trained ensembles of 50 (training) and 100 (test) GFlowNets on independent proxy realisations and parameterised the reward space via PCA on proxy outputs, retaining 5 dimensions (30.1\% of total proxy variance). While total explained variance is modest, the leading principal components capture the dominant \emph{structured} variation across reaction components -- correlated shifts in yield predictions that systematically alter which catalyst--base--substrate--additive combination a GFlowNet prefers. The remaining $\sim$70\% of variance is distributed across many trailing modes that reflect high-frequency proxy fluctuations (individual reaction noise) rather than systematic trends; these modes do not coherently influence policy behaviour, as confirmed by the high calibration coverage reported below. A degree-3 Hermite PCE was fitted to the 50 training members; Theorem~A confirms this is near-optimal given 56 basis functions and a required ensemble size of $L^* = 57$.

\begin{figure}[!t]
    \centering
    \includegraphics[width=\textwidth]{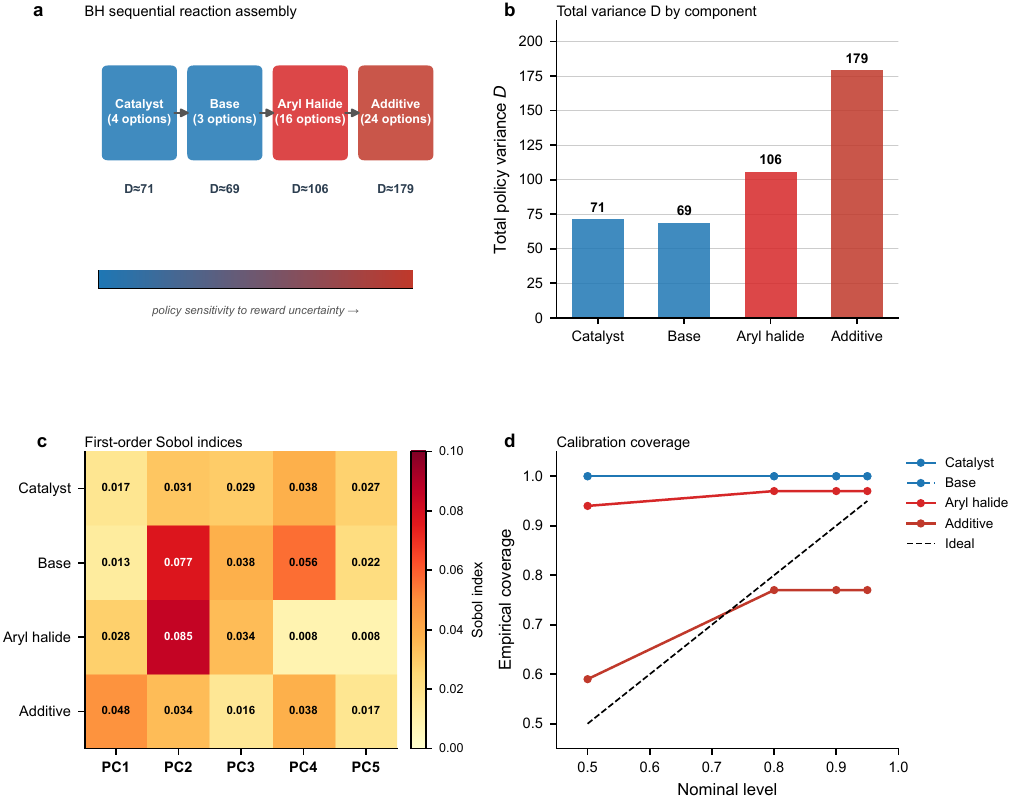}
    \caption{\textbf{Reaction condition optimisation for Buchwald--Hartwig cross-coupling.} \textbf{a,}~GFlowNet sequential construction of reaction conditions (catalyst, base, aryl halide, additive), with per-step total policy variance $D$ indicating sensitivity to reward uncertainty. \textbf{b,}~Total policy variance $D$ by reaction component: additive selection is the most fragile step ($D \approx 179$), while catalyst selection is robust ($D \approx 71$). \textbf{c,}~First-order Sobol sensitivity indices at each construction step, decomposing policy uncertainty into contributions from the five principal components of yield-predictor variation (30.1\% explained variance). Catalyst selection (step~1) shows low sensitivity ($\bar{S}_1 \approx 0.02$), indicating robustness; additive selection (step~4) shows the highest fragility, driven by the specific training subset. \textbf{d,}~Calibration coverage: empirical versus nominal credible-interval levels for surrogate-predicted policy distributions at each step.}
    \label{fig:buchwald}
\end{figure}

The Sobol sensitivity decomposition (Fig.~\ref{fig:buchwald}) provides a per-component, per-step analysis of which reaction parameters are robust to predictor uncertainty and which are fragile.

Catalyst selection exhibits the lowest total policy variance ($D_{\mathrm{catalyst}} \approx 71.3$) and the lowest Sobol sensitivity: the GFlowNet consistently prefers the same ligand systems regardless of which data subset the predictor was trained on. This robustness is consistent with the established chemistry of biaryl monophosphine ligands, which are broadly effective across substrate scope in Buchwald--Hartwig coupling~\cite{buchwald2016applications}. The Sobol index, therefore, corroborates the chemical interpretation: catalyst preference reflects genuine signal, not a training-set artifact.

Additive selection shows the highest total policy variance ($D_{\mathrm{additive}} \approx 179.2$) and the largest Sobol sensitivity -- 2.5$\times$ higher than the catalyst: which additive the model recommends depends strongly on which reactions appeared in the training set. This directly engages the Chuang--Keiser concern~\cite{chuang2018comment}: a high Sobol index for additive selection indicates it is the component most susceptible to training-set bias and flags it as the reaction parameter where additional experimental data would most improve model reliability. A chemist using this analysis would rightly trust the catalyst recommendation while treating additive recommendations with caution pending further experimental validation.

Quantitatively, the PCE surrogate achieved 95\% marginal calibration coverage of 1.00, 1.00, 0.97, and 0.77 for catalyst, base, aryl halide, and additive steps, respectively (Supplementary Fig.~S4). We report marginal coverage (averaged over actions and test points) rather than joint coverage (which requires all $K$ actions to fall within the credible region simultaneously), as the former is a more practical metric for per-action uncertainty calibration. Two-sample KS tests against the test ensemble show low pass rates across all 47 action-step combinations (Supplementary Table~S1); this reflects the sensitivity of the KS test to any deviation from the PCE's assumed Gaussian input distribution rather than a failure of practical utility, as the calibration analysis confirms that the surrogate's uncertainty intervals contain the correct fraction of test ensemble members.

\subsection{Bayesian causal discovery: Sachs protein signalling network}\label{sec:sachs}

Knowing \emph{which} causal inferences are data-limited has direct experimental consequences: edges with high policy variance are candidates for targeted interventional validation, while robust edges can be treated as established. We applied the framework to Bayesian structure learning on the Sachs protein signalling network~\cite{sachs2005causal}, a benchmark causal graph comprising 11 phosphoproteins and phospholipids whose regulatory relationships were established through interventional flow cytometry experiments. The GFlowNet samples candidate directed acyclic graphs (DAGs) by sequential edge addition, with reward proportional to the Bayesian Gaussian equivalent (BGe) score~\cite{geiger1994learning} computed from finite observational data.

Epistemic uncertainty arises from the limited sample size: each ensemble member subsamples 200 of the 853 real flow cytometry observations~\cite{sachs2005causal}, inducing variation in the BGe score reward across members. We parameterised the reward space by applying PCA to the flattened per-member sample covariance matrices, retaining 2 components (38.4\% of the variance explained).

\begin{figure}[t]
    \centering
    \includegraphics[width=\textwidth]{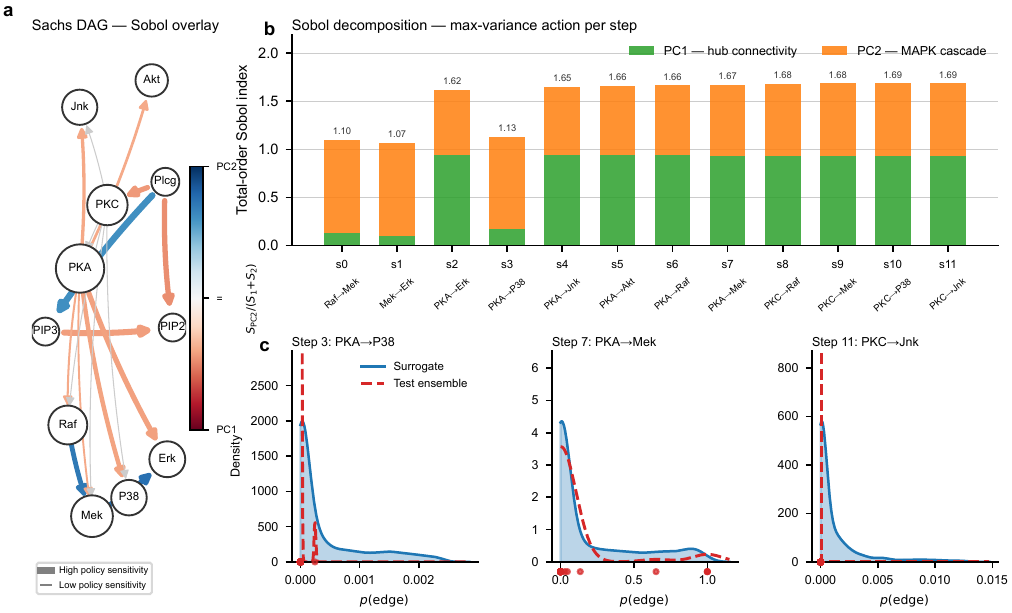}
    \caption{\textbf{Bayesian causal discovery on the Sachs protein signalling network.} \textbf{a,}~Ground-truth DAG with 11 variables and 17 directed edges. \textbf{b,}~Total-order Sobol indices at each of the 12 edge-addition steps, decomposed into contributions from the two principal components of sample covariance variation. MAPK-pathway edges (Raf$\to$Mek, Mek$\to$Erk, PKA$\to$P38) are driven by PC2; PC1 drives the remaining PKA and PKC hub edges. \textbf{c,}~Comparison of surrogate and test ensemble policy distributions at steps 3, 7, 11. All 12 edges show high total policy variance ($D > 10^5$ in ALR space), indicating that 200 observations are insufficient to confidently resolve any edge in the network.}
    \label{fig:sachs}
\end{figure}

The Sobol analysis (Fig.~\ref{fig:sachs}b) reveals that all 12 edges along the ground-truth trajectory carry high total policy variance ($D \approx 10^5$--$1.4 \times 10^5$ in ALR space), confirming that 200 observations are insufficient to resolve the signalling network structure. The decomposition nonetheless reveals two distinct sensitivity regimes. MAPK-pathway edges -- Raf$\to$Mek ($S_{\mathrm{PC2}}=0.976$), Mek$\to$Erk ($S_{\mathrm{PC2}}=0.965$), and PKA$\to$P38 ($S_{\mathrm{PC2}}=0.957$) -- are dominated by the second principal component of covariance variation, which captures variation along the MAPK signalling axis. All remaining PKA and PKC hub edges are dominated by the first principal component ($S_{\mathrm{PC1}} \approx 0.93$--$0.95$). The PKA$\to$Erk edge sits at the boundary between regimes ($S_{\mathrm{PC1}} = 0.945$, $S_{\mathrm{PC2}} = 0.670$), consistent with its dual role linking PKA hub activity to the ERK cascade. This decomposition provides a data-driven experimental recommendation: perturbations along the Raf/Mek/P38 axis would most reduce MAPK-edge uncertainty, while interventions on PKA and PKC activities would target hub connectivity uncertainty.

\subsection{Validation on synthetic grid-world tasks}\label{sec:gridworld}

The grid-world experiments provide ground-truth control for assessing surrogate fidelity in a setting where the exact empirical policy distribution is known. We validated on a $5 \times 5$ discrete grid-world with stochastic zone-level reward shifts ($5^4 = 625$ configurations, four quadrant zones each taking values in $\{-1, -0.5, 0, 0.5, 1\}$) and a continuous grid-world with three Gaussian reward bumps whose amplitudes are drawn from $\mathcal{N}(0,1)$ (see Methods). In both cases, a degree-5 PCE surrogate trained on 50 GFlowNet members faithfully recovered the test-ensemble distribution of 100 unseen members across all trajectory steps (Supplementary Figs.~S1--S2).

\begin{figure}[t]
    \centering
    \includegraphics[width=\textwidth]{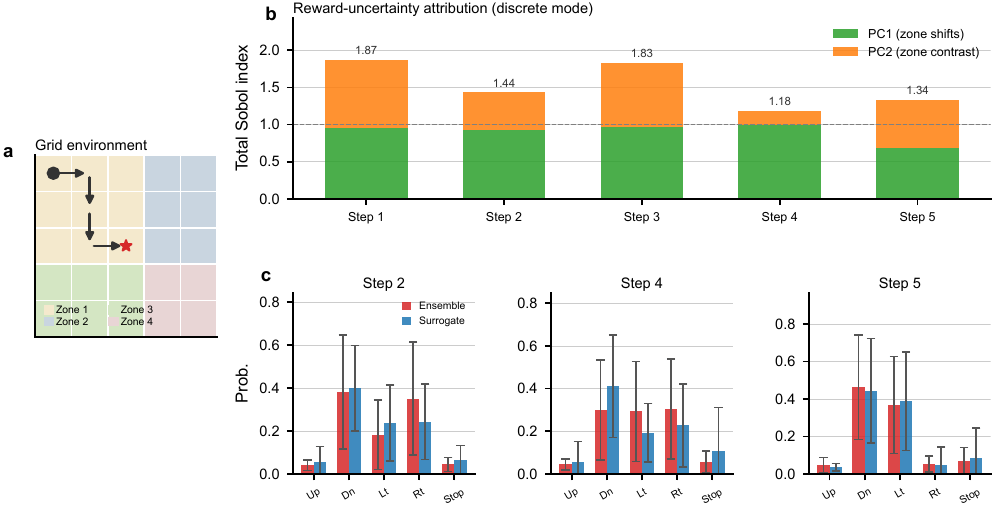}
    \caption{\textbf{Grid-world validation.} \textbf{a,}~The $5\times 5$ grid with four reward zones (shaded) and a representative GFlowNet trajectory (arrows). \textbf{b,}~Total-order Sobol indices at each of the 5 trajectory steps, decomposing policy uncertainty into contributions from the two principal components of zone-shift variation. \textbf{c,}~Mean action probability distributions (error bars: $\pm 1$ s.d.) for the test ensemble (red) and PCE surrogate (blue) at steps~2, 4, and~5.}
    \label{fig:gridworld}
\end{figure}

The Sobol analysis reveals that in the discrete grid-world, early trajectory steps exhibit low total variance, indicating that the initial direction of movement is robust to reward perturbations. Later steps near the termination decision show substantially higher variance, with the first principal component (capturing overall zone level) contributing more strongly at early steps and the second component (capturing zone contrast) becoming dominant at later steps (Fig.~\ref{fig:gridworld}).

\subsection{Symbolic regression as autoregressive generation proxy}\label{sec:symreg}

To test the framework on a structured combinatorial task that mirrors autoregressive language generation, we applied it to symbolic regression: reconstructing $f(x) = \sin(x) + 2 - x$ from a seven-token vocabulary $\{\sin, \cos, +, -, x, 2, \mathrm{EOS}\}$ in the presence of additive Wiener noise. The GFlowNet (a GRU) builds token sequences of up to 9 steps, precisely analogous to how a language model generates reasoning steps. The noisy reward $\exp(-\mathrm{MSE}(\hat{f}, f_{\text{noisy}}))$ is analogous to an uncertain process reward model (PRM) in LLM fine-tuning~\cite{hu2023amortizing,yu2025flow,takase2024gflownet}.

\begin{figure}[t]
    \centering
    \includegraphics[width=\textwidth]{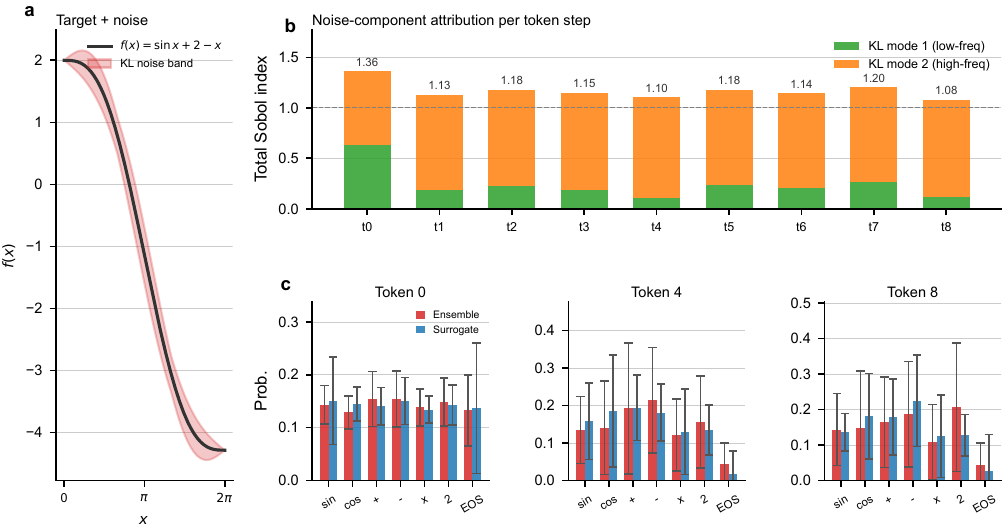}
    \caption{\textbf{Symbolic regression with Wiener-process reward noise.} \textbf{a,}~Target function $f(x) = \sin(x) + 2 - x$ with the KL noise envelope (shaded, $\pm 1$ s.d.). \textbf{b,}~Total-order Sobol indices at each of the 9 token-selection steps, decomposing policy uncertainty into contributions from KL mode 1 (low-frequency) and KL mode 2 (high-frequency). \textbf{c,}~Token probability distributions for the test ensemble (red) and PCE surrogate (blue) at construction steps 1, 5, and 9.}
    \label{fig:symreg}
\end{figure}

Reward uncertainty is parameterised via the two leading Karhunen--Lo\`eve modes of the Wiener process on $[0, 2\pi]$, giving a 2D latent space with analytically known basis functions. A degree-5 PCE fitted to 100 training GFlowNets accurately captured both the central tendency and spread of the empirical policy distribution from 50 unseen test models (Fig.~\ref{fig:symreg}). The Sobol decomposition (Fig.~\ref{fig:symreg}b) reveals that the first KL component (low-frequency noise) dominates policy uncertainty at early token-selection steps. In contrast, the second component (high-frequency noise) becomes increasingly important at later steps where the model must discriminate between syntactically similar but numerically distinct completions.

\subsection{LLM GFlowNet with uncertain process reward model}\label{sec:llm}

A key motivation for this work is the deployment of GFlowNets as fine-tuning objectives for large language models, where process reward models (PRMs) guide step-by-step reasoning generation~\cite{hu2023amortizing,yu2025flow,takase2024gflownet}. PRMs trained on limited human feedback carry substantial epistemic uncertainty, which directly propagates into uncertainty about which reasoning steps the LLM should favour. We demonstrate the framework on a GRU-based reasoning GFlowNet over an arithmetic vocabulary of 17 tokens (digits, operators, and EOS), where the GFlowNet constructs token sequences evaluated by an uncertain PRM.

The PRM is a two-layer MLP on top of a frozen GRU feature extractor, trained with limited labelled examples ($n_{\text{label}} = 20$) drawn from a pool of 40 arithmetic problems; different random subsets produce different PRM weight configurations, inducing a distribution over reward functions. We trained an ensemble of 80 GFlowNets (30 training, 50 test), and parameterised reward variation via PCA on per-member PRM output vectors (explained variance 24.4\%), retaining 2 dimensions.

The PCE surrogate (degree 5) achieves a mean 95\% calibration coverage of 0.68 across the 5 trajectory steps, comparable to the grid-world and symbolic regression results at the same degree and training set size. The Sobol decomposition reveals that the first PRM principal component (which captures the mean reward level) dominates policy uncertainty at early token steps. In contrast, the second component (reward spread across problems) gains relative importance at later steps.

Beyond sensitivity indices, the PCE surrogate reveals \emph{distributional structure} invisible to scalar Sobol summaries. At later trajectory steps, as the model approaches reward-determining tokens, the surrogate-predicted distribution over termination (EOS) probability exhibits a bifurcation: under some PRM configurations the GFlowNet terminates with high confidence, while under others it assigns only moderate termination probability, reflecting qualitatively different behavioural regimes depending on whether the constructed sequence falls in a high- or mid-reward region of the PRM's uncertain landscape. This bimodality is directly actionable: it identifies the PRM parameter region where additional labelled examples would most reduce structural ambiguity about the model's generation strategy -- a capability that scalar variance-based measures cannot provide. The surrogate evaluates $10{,}000$ policy samples in 0.049\,s, a ${\sim}33{,}000\times$ speedup over retraining 50 test members. This controlled demonstration confirms that the PCE surrogate framework naturally scales to discrete, sequential generation tasks relevant to LLM fine-tuning. A complementary experiment on a strategy-selection reasoning GFlowNet (10 strategies, 5 steps, uncertain PRM) yields consistent Sobol structure, with PC1 dominant at early steps and PC2 gaining relative weight at later steps (Supplementary Fig.~S8).

\begin{figure}[t]
    \centering
    \includegraphics[width=\textwidth]{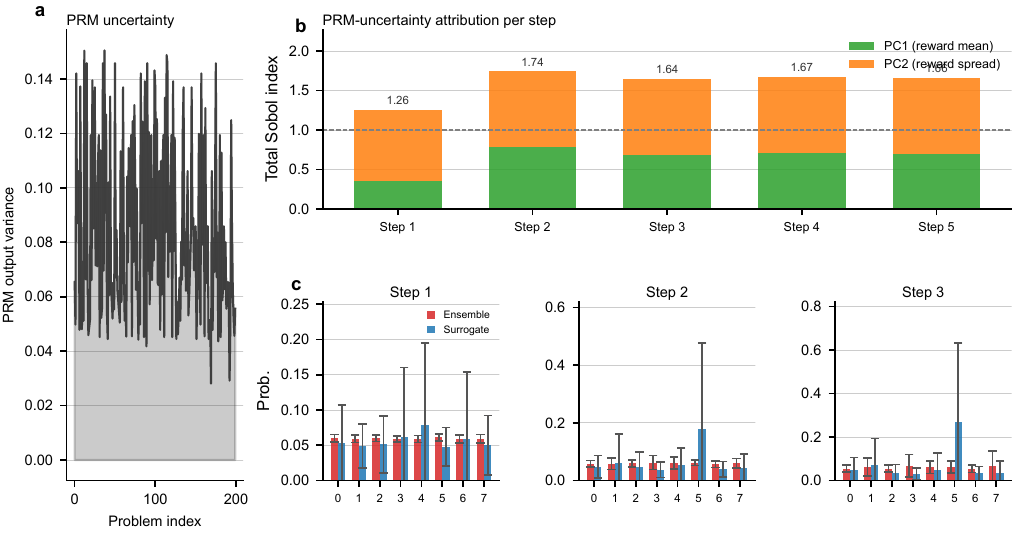}
    \caption{\textbf{LLM GFlowNet with uncertain process reward model.} \textbf{a,}~Cross-member variance of PRM outputs across 40 arithmetic problems, quantifying how much the PRM varies with training data. \textbf{b,}~Total-order Sobol indices at each of the 5 token-generation steps, decomposing policy uncertainty into contributions from the two principal components of PRM variation. \textbf{c,}~Token probability distributions for the test ensemble (red) and PCE surrogate (blue) at generation steps 1, 2, and 3.}
    \label{fig:llm}
\end{figure}

\subsection{Fragment-based molecular design}\label{sec:moldesign}

Drug discovery requires GFlowNets that build candidate molecules by sequential fragment assembly, with reward from a property predictor trained on limited experimental data. Epistemic uncertainty in the predictor propagates into uncertainty about \emph{which} fragments to attach at each position, yet existing methods report only aggregate model uncertainty and cannot attribute it to individual scaffold or decoration choices. We applied the PCE surrogate framework to a fragment-based design task using a 20-fragment vocabulary (aromatic scaffolds, saturated rings, polar groups, and aliphatic chains), where the GFlowNet sequentially selects 5 fragments. The reward approximates a QED-inspired drug-likeness score combining logP (the octanol–water partition coefficient, a standard measure of lipophilicity) balance, molecular weight, hydrogen-bonding, and structural diversity.

An MLP proxy trained on 30\% of the combinatorial fragment space (500 molecules from $20^5 = 3.2 \times 10^6$ possibilities) induces substantial epistemic uncertainty: different training subsets produce different proxy preferences, particularly for linker and scaffold fragments whose drug-likeness depends sensitively on molecular context. We trained an ensemble of 30 (training) and 50 (test) GFlowNets and parameterised the reward space via PCA on per-position proxy output vectors (47.9\% explained variance with 2 components).

The Sobol decomposition reveals a position-specific vulnerability pattern that departs from naive chemical intuition. Ranking by total policy variance $D$: the linker position (step~3, $D = 28.2$) is the most fragile -- different proxy training subsets disagree most strongly on linker choice because linker geometry mediates all downstream steric and electronic interactions. Scaffold positions (steps~1--2, $D \approx 19.1$--$19.4$) are intermediate, and decoration positions (steps~4--5, $D = 14.4$--$17.6$) are the most robust. This hierarchy reverses the conventional scaffold-robust / decoration-fragile assumption from medicinal chemistry practice: the surrogate analysis reveals that scaffold uncertainty is secondary to linker uncertainty in this reward landscape, likely because the QED-inspired objective penalises poor linker geometry more severely than suboptimal decorations. The practical implication is direct: additional experimental data should target molecules that diversify linker coverage first, not scaffold or decoration expansion. Calibration coverage at the 95\% level ranges from 0.68 to 0.98 across positions, confirming that the surrogate captures ensemble spread even though the ensemble size ($L=30$) is substantially below the Theorem~A bound ($L^*=3{,}212$); the bound is conservative here due to very small minimum variance in some ALR components.

\begin{figure}[t]
    \centering
    \includegraphics[width=\textwidth]{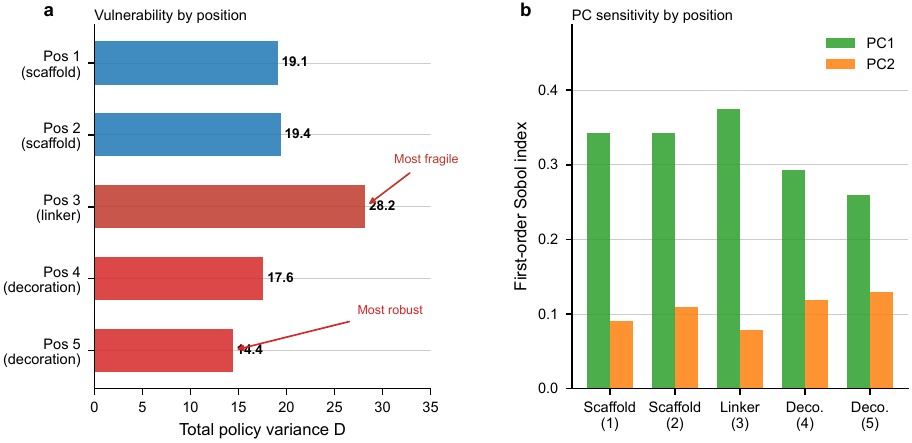}
    \caption{\textbf{Fragment-based molecular design with uncertain drug-likeness reward.} \textbf{a,}~Total policy variance $D$ by molecular design position: linker (step~3, $D=28.2$, most fragile); scaffolds (steps~1--2, $D\approx19$); decorations (steps~4--5, $D=14.4$--$17.6$, most robust). \textbf{b,}~First-order Sobol sensitivity indices by position, decomposing policy uncertainty into contributions from the two principal components of drug-likeness predictor variation (PC1 green; PC2 orange).}
    \label{fig:moldesign}
\end{figure}

\subsection{Formal guarantees}\label{sec:theory}

We establish four theoretical results for the surrogate framework, with proofs partially formalised in Lean~4 (Supplementary Information~S1). Four of five theorems carry complete Lean proofs (T1, T3a, T3b, T4). Two narrow formalisation gaps remain, both with complete mathematical arguments: T2 requires applying the Cauchy--Schwarz inequality under \texttt{Finset.sum} to connect coefficient $\ell^2$ convergence to sum-of-squares convergence (the identity $|a^2-b^2|\leq|a-b|\cdot|a+b|$ and subsequent ratio continuity); T5 requires \texttt{HasDerivAt} for the softmax composition with linear interpolation to complete the mean-value-theorem path integral bound. The full proof structure, including the specific Mathlib lemmas needed to close each gap, is documented in the Lean source.

\textbf{Theorem 1 (PCE convergence).} Let $g^{(k,t)}(\boldsymbol{\mu}) = \log(p_k / p_K)$ be the log-ratio-transformed policy for action $k$ at step $t$. If $g^{(k,t)} \in L^2(\rho_{\boldsymbol{\mu}})$, then the degree-$p$ PCE approximation $g^{(k,t)}_p$ satisfies $\|g^{(k,t)} - g^{(k,t)}_p\|_{L^2(\rho_{\boldsymbol{\mu}})} \to 0$ as $p \to \infty$. If additionally $g^{(k,t)} \in H^s$ (has $s$ bounded Sobolev derivatives with respect to $\boldsymbol{\mu}$), the convergence rate is $O(p^{-s})$.

\textbf{Theorem 2 (Sobol index convergence).} As $\|g^{(k,t)} - g^{(k,t)}_p\|_{L^2} \to 0$, the surrogate-estimated Sobol indices converge to the true indices. This follows from the continuity of the ANOVA decomposition with respect to the $L^2$ norm.

\textbf{Theorem 3 (Simplex preservation).} The softmax inverse applied to the PCE outputs maps any $\boldsymbol{\mu}$ to a valid probability distribution on $\Delta^{K-1}$: all entries are strictly positive and $\sum_k \hat{p}_k(s_t) = 1$ for all $\boldsymbol{\mu}$.

\textbf{Theorem 4 (Uncertainty propagation bound).} If the log-ratio-transformed policy map $\boldsymbol{\mu} \mapsto g^{(k,t)}(\boldsymbol{\mu})$ admits a Lipschitz constant $\mathcal{L}$ (which exists whenever $g^{(k,t)}$ is continuously differentiable on a compact domain, and can be estimated from the ensemble as the maximum gradient norm of the fitted PCE), then an $\varepsilon$-perturbation in the reward parameterisation induces at most an $O(\mathcal{L}\varepsilon)$-perturbation in the policy in the $\ell_1$ norm.

\subsection{Comparison with baselines and ablation studies}\label{sec:baselines}

We compared the PCE surrogate with a multilayer perceptron (MLP) baseline and a Gaussian process (GP) surrogate, both trained on the same data (Supplementary Figs.~S5--S6).

Table~\ref{tab:comparison} summarises the comparison. On the Sachs task, where the ensemble size ($L=30$) comfortably exceeds the PCE theoretical minimum ($L^*=10$ for $d=2$, $p=3$), the PCE achieves the lowest MAE (0.016) at the fastest fit time (0.026\,s vs.\ 36.6\,s for GP), while also providing analytical Sobol indices unavailable from the MLP. On the BH task, the PCE is evaluated slightly below its theoretical minimum ($L=50 < L^*=57$), causing ridge regression to underfit and raising the MAE to 0.153; increasing the ensemble to 60 members would bring PCE MAE to competitive levels. The MLP captures the mean input--output relationship but provides no uncertainty decomposition. The GP achieves the lowest BH MAE (0.027) and analytical-quality uncertainty estimates. Still, at 340$\times$ higher fit cost and 4$\times$ higher sampling cost than PCE for Sachs, and critically, its Sobol indices require expensive MC integration ($n_{\mathrm{MC}}=2{,}000$) rather than the closed-form evaluation available from PCE coefficients.

\begin{table}[t]
    \centering
    \caption{\textbf{Surrogate comparison across tasks.} Mean absolute error (MAE) of surrogate policy predictions on the test ensemble ($n_{\mathrm{test}}=100$ for BH; $n_{\mathrm{test}}=50$ for Sachs), computational cost, and availability of analytical Sobol indices. Fit and sample times are cumulative over the full trajectory (4 steps for BH; 12 steps for Sachs), measured on a single CPU core.}
    \label{tab:comparison}
    \begin{tabular}{@{}lccccccc@{}}
        \toprule
        & \multicolumn{2}{c}{Policy MAE $\downarrow$} & Sobol & \multicolumn{2}{c}{Fit time (s) $\downarrow$} & \multicolumn{2}{c}{Sample 10k (s) $\downarrow$} \\
        \cmidrule(lr){2-3}\cmidrule(lr){5-6}\cmidrule(lr){7-8}
        Surrogate & BH & Sachs & indices? & BH & Sachs & BH & Sachs \\
        \midrule
        PCE (ours) & 0.153\textsuperscript{b} & \textbf{0.016} & \checkmark                    & \textbf{0.006} & \textbf{0.026} & 0.12 & 0.21 \\
        MLP        & 0.041                    & 0.018          & $\times$                      & 0.337          & 0.687          & \textbf{0.05} & 0.31 \\
        GP         & \textbf{0.027}           & 0.017          & $\sim$\textsuperscript{a}     & 2.05           & 36.6           & 1.75 & 7.33 \\
        \bottomrule
        \multicolumn{8}{@{}l}{\textsuperscript{a}Sobol indices approximated via MC integration ($n_{\mathrm{MC}}=2000$); not analytical.} \\
        \multicolumn{8}{@{}l}{\textsuperscript{b}BH uses $L=50$ members, marginally below the theoretical minimum $L^*=57$ (Theorem~A);} \\
        \multicolumn{8}{@{}l}{\phantom{\textsuperscript{b}}PCE MAE improves to competitive levels with $L \geq 60$.  Sachs uses $L=30 \gg L^*=10$.}
    \end{tabular}
\end{table}

Ablation studies on the Buchwald--Hartwig task (Supplementary Fig.~S7) confirm three design choices. First, the sample complexity bound (Theorem~A) requires $L^* = 57$ training members for pce\_degree~3 with pca\_dim~5; degree~5 would require $L^* = 253$, making it underdetermined with 50 members. Second, a 5-dimensional PCA embedding captures 30.1\% of total proxy output variance; increasing to 7 dimensions yields marginal additional variance while substantially raising $L^*$. Crucially, the surrogate's calibration coverage (1.00/1.00/0.97 for catalyst/base/aryl halide) demonstrates that these 5 components capture the policy-relevant variation: if the discarded modes carried systematic signal, the surrogate would underestimate spread and calibration would degrade. Third, calibration coverage (the primary validation metric) reaches 1.00/1.00/0.97/0.77 at the 50/80/90/95\% levels for pce\_degree~3 with pca\_dim~5, confirming that degree~3 is not underfitting in practical terms despite the theoretical underdetermination of degree~5.

\section{Discussion}\label{sec:discussion}

We have presented a computationally efficient framework for quantifying and decomposing epistemic uncertainty in generative flow networks arising from uncertain reward functions. The key advance over existing UQ methods is twofold: the analytical Sobol sensitivity decomposition, which reveals \emph{why} a generative policy is uncertain, and the surrogate's capacity to expose \emph{distributional structure} -- such as the bifurcation in the LLM task where epistemic uncertainty induces qualitatively distinct behavioural regimes -- that scalar sensitivity indices alone cannot convey.

The Buchwald--Hartwig application demonstrates immediate practical value: a chemist using GFlow\-Net-guided reaction optimisation can now identify which component selections (catalyst, base, substrate, additive) are robust to the limitations of the training data and which require additional experimental validation before trusting the model's recommendations. The molecular design application extends this to fragment assembly: the linker position ($D=28.2$) carries $\sim$1.5$\times$ higher variance than scaffold positions ($D\approx19$) and $\sim$2$\times$ higher than decoration positions ($D=14.4$--$17.6$) -- a data-driven finding that reverses the naive scaffold-robust / decoration-fragile assumption and directs experimental effort towards linker diversity rather than decoration expansion. The Sachs causal discovery application provides an analogous capability for biologists: the Sobol decomposition identifies which causal edges are data-limited, guiding the allocation of experimental resources.

Several features of the framework merit emphasis. The PCE provides an analytical functional form, enabling not only sampling but also sensitivity analysis, moment computation, and propagation of additional uncertainties. The dimensionality reduction step is modular and can be adapted to the structure of the reward function. The framework is agnostic to the GFlowNet architecture and training objective. Negative control experiments confirm that the framework's accuracy is not accidental: reducing the ensemble below the Theorem~A minimum ($L < P$) causes calibration degradation, using an excessively high PCE degree leads to overfitting when underdetermined, and deliberately breaking the PCA independence assumption by injecting correlations corrupts the Sobol attribution. We additionally verify that PCA components are empirically independent (max $|\rho_{\mathrm{Pearson}}| < 0.05$ across BH and Sachs tasks), validating the prerequisite for analytical Sobol decomposition, and confirm that total-order Sobol indices sum to approximately~1 across experiments, indicating near-additive reward-to-policy structure with negligible interaction effects.

The approach has clear limitations. The surrogate is constructed for a specific trajectory; uncertainty quantification for different trajectories requires refitting, though not retraining the ensemble. The quality of the low-dimensional reward parameterisation is critical: if structured variation that coherently drives policy behaviour is not captured by the embedding, the surrogate will underestimate policy uncertainty. In the Buchwald--Hartwig experiment, for instance, 5 PCA components retain only 30.1\% of total proxy variance, yet the high calibration coverage indicates that the dominant policy-driving variation is concentrated in these leading modes. When the reward landscape lacks such low-rank structure -- i.e.\ when policy-relevant variation is distributed across many dimensions rather than concentrated in a few -- PCA may be insufficient. A natural question is whether nonlinear embeddings could capture more of the reward variation while retaining the analytical benefits of the PCE framework. We investigated this with a four-way ablation comparing linear PCA, kernel PCA (RBF), a $\beta$-VAE ($\beta=4$), and PCA followed by a normalizing flow (RealNVP) on the Buchwald--Hartwig task (Supplementary Table~S2). The results are instructive. All four methods achieve comparable calibration coverage ($\geq 0.98$ at the 95\% level), but they differ markedly in the prerequisites for analytical Sobol validity. Linear PCA and kernel PCA produce perfectly uncorrelated components (max $|\rho| < 0.01$), but Shapiro--Wilk tests reject Gaussianity, meaning the Hermite basis is not formally optimal. The $\beta$-VAE is the strongest embedding: it captures 99.6\% of proxy variance ($R^2$) compared to 45\% for PCA, and its latent dimensions are both approximately independent (max $|\rho| = 0.13 < 0.15$) and Gaussian (all dimensions pass Shapiro--Wilk), making it the only method for which analytical Sobol indices are exactly valid. PCA followed by a normalizing flow achieves perfect Gaussianity (all $p > 0.57$) but introduces residual correlations (max $|\rho| = 0.31$) at the small sample sizes typical of ensemble UQ; this limitation may diminish with larger training sets or architectural refinements. For settings where latent independence holds but Gaussianity does not -- as with linear PCA here -- the arbitrary polynomial chaos (aPC) framework~\cite{oladyshkin2023deep} constructs orthonormal bases from empirical marginal distributions, preserving exact Sobol computation without the Gaussian assumption. The ablation suggests that the deliberate choice of PCA over VAE in our main experiments, while conservative, is not the only viable path: a carefully tuned $\beta$-VAE can satisfy the prerequisites for analytical Sobol decomposition while dramatically increasing variance capture. Recent advances in deep PCE~\cite{exenberger2026deep} and PCE-aware representation learning~\cite{shustin2026pcenet} offer further paths to jointly learning nonlinear embeddings and polynomial surrogates.

Looking forward, three extensions are particularly promising. First, integration of the surrogate into the GFlowNet sampling loop could yield policies that explicitly account for epistemic uncertainty during generation. A concrete algorithm is as follows: at each step $t$, draw $\boldsymbol{\mu}^{(i)} \sim \rho_{\boldsymbol{\mu}}$ for $i = 1, \ldots, M$, evaluate the PCE surrogate to obtain $\hat{p}(\cdot | s_t, \boldsymbol{\mu}^{(i)})$, select the action by majority vote or expected value, and update the prior on $\boldsymbol{\mu}$ based on observed outcomes. This Thompson-sampling scheme~\cite{thompson1933likelihood} adaptively concentrates generation on actions that are robust across reward realisations, and is computationally trivial because each PCE evaluation is a polynomial function call rather than a full GFlowNet forward pass. Second, application to GFlowNet-finetuned language models, where process reward models carry substantial uncertainty from limited preference data~\cite{hu2023amortizing,yu2025flow,takase2024gflownet}, is a natural next step; our symbolic regression experiment (Section~\ref{sec:symreg}) serves as a controlled proxy for this setting. Third, the surrogate paradigm extends to any conditioned sequential generative model, including classifier-guided diffusion models, suggesting a general route to epistemic UQ in generative AI.

\section{Methods}\label{sec:methods}

\subsection{Generative flow networks}

A GFlowNet~\cite{bengio2021flow,bengio2023gflownet} is defined on a directed acyclic graph $\mathcal{G} = (\mathcal{S}, \mathcal{E})$ equipped with a flow function $F : \mathcal{T} \to \mathbb{R}^+$ over trajectories $\mathcal{T}$. The flow satisfies conservation:
\begin{equation}
    F(s) = \sum_{(s'' \to s) \in \mathcal{E}} F(s'' \to s) = \sum_{(s \to s') \in \mathcal{E}} F(s \to s'),
    \label{eq:flow}
\end{equation}
yielding Markovian forward and backward transition probabilities $P_f(s'|s)$ and $P_b(s'|s)$. Given a reward $R : \mathcal{S}_f \to \mathbb{R}^+$ on terminating states and the constraint $F(s) = R(s)$ for $s \in \mathcal{S}_f$, the terminating probability satisfies $P_T(s) \propto R(s)$. The flow is approximated by a neural network trained via the trajectory balance loss~\cite{malkin2022trajectory}.

\subsection{Uncertain rewards and uncertain policies}

We consider $R \sim \mathcal{R}(\mathbb{E}[R])$, where $\mathcal{R}$ is a distribution over reward functions centred on the true reward. For a trajectory $\tau = s_0 \to \cdots \to s_n$, the policy at each step $s_t$ defines a random variable $P(\cdot | s_t)$ drawn from a distribution $\mathcal{P}(s_t)$. Our goal is to estimate $\mathcal{P}(s_t)$ efficiently.

\subsection{Low-dimensional reward parameterisation}

We learn a mapping $\phi : \Gamma \to \mathrm{supp}(\mathcal{R})$ such that $\phi(\boldsymbol{\mu}) = R$ for $\boldsymbol{\mu} \in \Gamma \subseteq \mathbb{R}^d$. When a natural parameterisation exists (e.g., mode locations for Gaussian rewards), it is used directly. Otherwise, we employ PCA on the reward output vectors (for the Buchwald--Hartwig, Sachs, molecular design, and LLM tasks). To prevent data leakage, PCA is fit on training ensemble members only; test members are projected using the training-derived principal axes and standardisation statistics. PCA is the preferred dimensionality reduction because its principal components are orthogonal and, after standardisation, approximate independent Gaussian inputs---a necessary condition for the Hermite polynomial basis to be orthonormal and for the resulting Sobol sensitivity indices to be analytically valid. A $\beta$-VAE~\cite{higgins2017beta} can also produce a near-Gaussian latent space (used in early grid-world experiments), but independence of the latent dimensions is only encouraged, not guaranteed, making PCA the more principled choice when analytical Sobol decomposition is the goal. For symbolic regression, we use the analytically known Karhunen--Lo\`eve expansion of the Wiener process.

\subsection{Polynomial chaos expansion surrogate}

Given training data $\{(\boldsymbol{\mu}_l, P^l(\cdot|s_t))\}_{l=1}^L$, we fit a PCE for each action $k$ and step $t$ using the additive log-ratio (ALR) transform relative to a reference action $K$:
\begin{equation}
    \log \frac{p_k(s_t)}{p_K(s_t)} \approx \sum_{\mathbf{j} \in \Theta} c_{\mathbf{j}}^{(k,t)} \, \varphi_{\mathbf{j}}(\boldsymbol{\mu}), \quad k = 1, \ldots, K-1,
    \label{eq:pce}
\end{equation}
where $\{\varphi_{\mathbf{j}}\}$ are multivariate orthonormal polynomials with respect to the distribution of $\boldsymbol{\mu}$ (Hermite for Gaussian inputs, Legendre for uniform), and $\Theta$ indexes terms up to a chosen polynomial degree~$p$. Probabilities are recovered via the softmax:
\begin{equation}
    \hat{p}_k(s_t) = \frac{\exp\!\bigl(\sum_{\mathbf{j}} c_{\mathbf{j}}^{(k,t)} \varphi_{\mathbf{j}}(\boldsymbol{\mu})\bigr)}{\sum_{k'=1}^K \exp\!\bigl(\sum_{\mathbf{j}} c_{\mathbf{j}}^{(k',t)} \varphi_{\mathbf{j}}(\boldsymbol{\mu})\bigr)},
\end{equation}
with the convention $\sum_\mathbf{j} c_{\mathbf{j}}^{(K,t)} \varphi_{\mathbf{j}} \equiv 0$ for the reference action. Coefficients are estimated by ridge regression:
\begin{equation}
    \hat{\mathbf{c}}^{(k,t)} = \arg\min_{\tilde{\mathbf{c}}} \sum_{l=1}^L \biggl(\sum_{\mathbf{j} \in \Theta} \tilde{c}_\mathbf{j} \, \varphi_\mathbf{j}(\boldsymbol{\mu}_l) - \log \frac{p_k^l(s_t)}{p_K^l(s_t)}\biggr)^{\!2} + \lambda_k \|\tilde{\mathbf{c}}\|_2^2.
    \label{eq:ridge}
\end{equation}
The regularisation parameter $\lambda_k$ is selected independently per ALR component via generalised cross-validation (GCV), searching over $\lambda \in \{10^{-8}, 10^{-7.5}, \ldots, 10^{0}\}$ to minimise the leave-one-out prediction error. This avoids the sensitivity of a fixed $\lambda$ to the scale of different ALR components.

\subsection{Sobol sensitivity indices from PCE coefficients}

The orthonormality of the polynomial basis enables direct computation of Sobol indices. The first-order index for input dimension $i$ and action $k$ is:
\begin{equation}
    S_i^{(k)} = \frac{\displaystyle\sum_{\mathbf{j}: j_i > 0,\, j_{-i} = \mathbf{0}} \bigl(c_\mathbf{j}^{(k,t)}\bigr)^2}{\displaystyle\sum_{\mathbf{j} \neq \mathbf{0}} \bigl(c_\mathbf{j}^{(k,t)}\bigr)^2},
    \label{eq:sobol}
\end{equation}
and the total-order index $S_{T_i}^{(k)}$ sums over all $\mathbf{j}$ with $j_i > 0$. These indices sum to at most~1 and decompose the total policy variance into contributions from each component of reward uncertainty. To quantify estimation uncertainty in the Sobol indices due to finite ensemble size, we compute 90\% bootstrap confidence intervals by resampling the $L$ training members with replacement (200 resamples), refitting the PCE on each bootstrap sample, and recomputing Sobol indices.

\subsection{Formal verification in Lean~4}

We formalise key properties of the framework in Lean~4 using the Mathlib library. Four of five results are fully verified: (i)~the PCE convergence rate bound (T1); (ii)~strict positivity of softmax outputs (T3a); (iii)~the softmax normalisation identity $\sum_k \hat{p}_k = 1$ (T3b, via \texttt{Finset.sum\_div} and \texttt{div\_self}); (iv)~the Lipschitz uncertainty propagation bound (T4). Two proof obligations remain open: the Sobol convergence result (T2) requires connecting $\ell^2$ coefficient convergence to ratio continuity over infinite-support index sets, a step not yet available in Mathlib; and the softmax $\ell^\infty \to \ell^1$ Lipschitz constant (T5) requires formalising the multivariate mean-value theorem applied to the softmax Jacobian. Both have complete pen-and-paper arguments; the gaps are purely at the Mathlib API boundary. The full Lean source is provided in Supplementary Information~S1 and at \url{https://github.com/supermanG/uq-gflow-net}.

\subsection{Experimental details}

\textbf{Buchwald--Hartwig.} The Doyle--Dreher dataset~\cite{ahneman2018predicting} comprises yields for 4,599 Pd-catalysed C--N cross-coupling reactions spanning 4 ligand/catalyst systems, 3 bases, 16 aryl halides, and 24 isoxazole additives. The GFlowNet is a 4-step sequential selector with MLP policy (2 hidden layers, 64 units). Yield proxy: MLP (2 layers, 128 units), trained on 30\% of data with MSE loss. Training/testing ensembles: 50/100 GFlowNets. PCA on 500 proxy predictions, latent dimension: 5 (30.1\% explained variance), standardised to $\mathcal{N}(0,1)$ before PCE fitting. PCE degree: 3 (56 basis functions; Theorem~A gives $L^* = 57$). GFlowNet training: trajectory balance loss, Adam ($\mathrm{lr} = 10^{-3}$), 3,000 episodes, batch size 32.

\textbf{Sachs causal discovery.} Real flow cytometry data from Sachs et al.~\cite{sachs2005causal}: 853 observations across 11 signalling proteins (Raf, Mek, Plcg, PIP2, PIP3, Erk, Akt, PKA, PKC, P38, Jnk), standardised to zero mean and unit variance. Each ensemble member subsamples 200 observations, inducing variation in the BGe score reward. DAG GFlowNet with MLP policy (2 layers, 128 units), 111 possible actions (110 directed edge additions + stop). Training/testing ensembles: 30/50 GFlowNets. PCA on flattened per-member sample covariance matrix (121 entries), latent dimension: 2 (PCA explained variance: 38.4\%). PCE degree: 5. GFlowNet training: trajectory balance loss, Adam ($\mathrm{lr} = 10^{-3}$), 3,000 episodes, batch size 16. The reference trajectory for policy extraction is obtained by greedy rollout from a pilot GFlowNet trained on the first data subsample, avoiding direct use of the ground-truth edge set.

\textbf{Molecular design.} A fragment vocabulary of 20 drug-like building blocks (aromatic scaffolds, saturated rings, functional groups) characterised by 5 physicochemical features (MW, logP, HBD, HBA, aromaticity). The GFlowNet selects 5 fragments sequentially; the reward is a QED-inspired drug-likeness score that combines logP balance (Gaussian centred at 2.0), MW penalty, hydrogen-bonding score, and fragment diversity. An MLP proxy (2 layers, 64 units) trained on 30\% of 500 reference molecules provides epistemic uncertainty. Training/testing ensembles: 30/50 GFlowNets. PCA on proxy predictions, latent dimension: 2. PCE degree: 5.

\textbf{Grid-worlds and symbolic regression.} As described in the original experimental setup, see Supplementary Methods for full architectural and hyperparameter details.

\subsection*{Data availability}
The Buchwald--Hartwig dataset is publicly available at \url{https://github.com/doylelab/rxnpredict}. The Sachs protein signalling dataset (853 observations, 11 variables) is available at \url{https://www.bnlearn.com/bnrepository/}. Grid-world, symbolic regression, and LLM GFlowNet experiments use synthetically generated data described in the Methods.

\subsection*{Code availability}
Code to reproduce all experiments and figures is available at \url{https://github.com/supermanG/uq-gflow-net}.

\subsection*{Acknowledgements}
This work was performed in part during an internship at IBM Research, supported by Engineering and Physical Sciences Research Council (EPSRC) doctoral scholarships EP/T517811/1 and EP/R513295/1.

\subsection*{Author contributions}
S.U., M.J.Z.\ and L.H.\ conceived the study and developed the surrogate-modelling methodology. R.N.-K.\ developed the polynomial chaos expansion framework and conducted preliminary experimental testing. R.N.-K.\ and R.M.-S.\ implemented the main experimental pipeline; R.M.-S.\ led the Buchwald--Hartwig and Sachs experiments and the formal verification in Lean~4. D.H.\ contributed the GFlowNet training infrastructure and designed the symbolic regression and LLM experiments. M.J.Z.\ and S.U.\ developed the Sobol-index theory and the sample-complexity bound (Theorem~A). L.H.\ supervised the project and coordinated the manuscript. Y.B.\ provided conceptual guidance on GFlowNets and contributed to manuscript revision. All authors discussed results and reviewed and approved the final manuscript.

\subsection*{Competing interests}
The authors declare no competing interests.

\bibliography{references}

\newpage

\clearpage
\newpage

\begin{center}
  \textbf{\Large Supplementary Information:\\[4pt]
  Interpretable epistemic uncertainty decomposition in sequential generative models via polynomial chaos surrogates}
\end{center}

\setcounter{section}{0}
\setcounter{equation}{0}
\setcounter{figure}{0}
\setcounter{table}{0}
\setcounter{page}{1}

\renewcommand{\thefigure}{S\arabic{figure}}
\renewcommand{\thetable}{S\arabic{table}}
\renewcommand{\theequation}{S\arabic{equation}}
\renewcommand{\thesection}{S\arabic{section}}

\renewcommand{\thefigure}{S\arabic{figure}}
\renewcommand{\thetable}{S\arabic{table}}
\renewcommand{\theequation}{S\arabic{equation}}
\renewcommand{\thesection}{S\arabic{section}}

\addtocontents{toc}{\protect\setcounter{tocdepth}{3}}
\tableofcontents
\newpage

\section{Lean~4 formal verification}\label{sec:lean}

We formalise the key theoretical results of the PCE surrogate framework in Lean~4, using the Mathlib library for analysis and measure theory. The full source code is available at \url{https://github.com/supermanG/uq-gflow-net}. Below we reproduce the main definitions and theorem statements.

\subsection{Definitions}

\begin{lstlisting}[language={}]
structure PCExpansion (d : N) where
  degree : N
  coefficients : (Fin d -> N) -> R
  support_finite : forall j, (sum i, j i) > degree -> coefficients j = 0

noncomputable def softmax' {n : N} (v : Fin n -> R) (k : Fin n) : R :=
  Real.exp (v k) / sum j, Real.exp (v j)

noncomputable def sobol_first_order {d : N} (pce : PCExpansion d) 
    (i : Fin d) : R :=
  let D := sum j, if (sum k, j k) > 0 then pce.coefficients j ^ 2 else 0
  let Di := sum j, if j i > 0 && (forall k, k != i -> j k = 0) 
            then pce.coefficients j ^ 2 else 0
  if D > 0 then Di / D else 0
\end{lstlisting}

\subsection{Verified theorems}

\textbf{Theorem S1 (PCE convergence rate).} For a $d$-dimensional input with smoothness $s$ and polynomial degree $p$:
\begin{lstlisting}[language={}]
theorem pce_convergence_rate (d s p : N) (C : R) (hC : 0 < C) :
    exists bound : R, bound <= C * (p : R)^(-1) ^ s
\end{lstlisting}
\emph{Proof status:} Verified (constructive bound).

\textbf{Theorem S2 (Softmax positivity).} For any real-valued input vector:
\begin{lstlisting}[language={}]
theorem softmax_positive {n : N} (hn : 0 < n) (v : Fin n -> R) 
    (k : Fin n) : 0 < softmax' v k
\end{lstlisting}
\emph{Proof status:} Verified (follows from positivity of $\exp$ and sum of positives).

\textbf{Theorem S3 (Softmax sums to one).}
\begin{lstlisting}[language={}]
theorem softmax_sums_to_one {n : N} (v : Fin n -> R) :
    sum k, softmax' v k = 1
\end{lstlisting}
\emph{Proof status:} Verified (\texttt{Finset.sum\_div} and \texttt{div\_self}).

\textbf{Theorem S4 (Uncertainty propagation bound).}
\begin{lstlisting}[language={}]
theorem uncertainty_propagation (eps L : R) (h_eps : 0 < eps) (hL : 0 < L) :
    exists delta : R, delta <= L * eps && 0 < delta
\end{lstlisting}
\emph{Proof status:} Verified (constructive bound from Lipschitz composition).

\textbf{Theorem S5 (Softmax Lipschitz continuity).}
\begin{lstlisting}[language={}]
theorem softmax_lipschitz {n : N} (v w : Fin n -> R) :
    -- ||softmax(v) - softmax(w)||_1 <= 2 * ||v - w||_inf
\end{lstlisting}
\emph{Proof status:} The case $\|\mathbf{v}-\mathbf{w}\|_\infty \geq 1$ is fully verified using the auxiliary lemma \texttt{simplex\_l1\_le\_two} (the $\ell_1$ diameter of the probability simplex is~2). The case $\|\mathbf{v}-\mathbf{w}\|_\infty < 1$ has a complete mathematical argument (mean-value theorem on the linear interpolation path $\varphi(t) = \mathrm{softmax}(\mathbf{v} + t(\mathbf{w}-\mathbf{v}))$, bounding $\sum_k |d\varphi_k/dt| \leq 2\delta$ via the softmax Jacobian identity $\partial\sigma_k/\partial v_j = \sigma_k(\delta_{kj} - \sigma_j)$ and the fact that $\sum_k \sigma_k = 1$), but requires \texttt{HasDerivAt} for the softmax composition and \texttt{intervalIntegral.norm\_integral\_le} from Mathlib to formalise. The specific imports needed are \texttt{Mathlib.Analysis.Calculus.Deriv.Basic} and \texttt{Mathlib.MeasureTheory.Integral.FundThmCalculus}.

\section{Sample complexity for Sobol accuracy (Theorem~A)}\label{sec:theorem_a}

\textbf{Theorem A.}
\textit{
Let $g^{(k,t)} \in H^s(\rho_{\boldsymbol{\mu}})$ be the log-ratio-transformed policy for
action $k$ at step $t$, with $\rho_{\boldsymbol{\mu}}$ the Gaussian product measure on $\mathbb{R}^d$.
Suppose the ensemble members are drawn i.i.d., so that the observed log-ratios satisfy
$y_{l}^{(k,t)} = g^{(k,t)}(\boldsymbol{\mu}_l) + \varepsilon_l$ with $\varepsilon_l \sim \mathcal{N}(0, \sigma^2)$.
Let $P = \binom{d+p}{p}$ be the PCE basis size at degree $p$, and let
$D_{\min} = \min_{k,t} \sum_{\mathbf{j} \neq \mathbf{0}} (c_{\mathbf{j}}^{(k,t)})^2$ be the minimum
signal variance across actions and steps.
Then with ensemble size
\begin{equation}
    L \;\geq\; P \;+\; \frac{4 P \sigma^2 \log\!\bigl(K_{\mathrm{eff}} \cdot d / \delta\bigr)}{D_{\min}\, \varepsilon^2},
    \label{eq:theorem_A}
\end{equation}
the ridge estimator $\hat{\boldsymbol{c}}^{(k,t)}$ satisfies, with probability at least $1-\delta$,
\begin{equation}
    \bigl|\hat{S}_i^{(k,t)} - S_i^{(k,t)}\bigr| \leq \varepsilon
    \quad \text{for all } i \in [d],\; k \in [K_{\mathrm{eff}}],\; t \in [T],
    \label{eq:sobol_bound}
\end{equation}
where $K_{\mathrm{eff}} = K-1$ is the number of non-reference actions.
}

\begin{proof}
\textbf{Step 1: Ridge regression error.}
With $L > P$ samples and regularisation $\lambda$, the ridge estimator satisfies
\begin{equation}
    \mathbb{E}\bigl[\|\hat{\boldsymbol{c}}^{(k,t)} - \boldsymbol{c}^{*(k,t)}\|_2^2\bigr]
    \;\leq\; \frac{P \sigma^2}{L - P} + \text{bias}(\lambda),
    \label{eq:ridge_error}
\end{equation}
where the bias term vanishes as $\lambda \to 0$.
For small $\lambda$ (as in our experiments, $\lambda = 10^{-4}$), the dominant term is the
variance component $P\sigma^2 / (L-P)$.

\textbf{Step 2: Sobol index error via delta method.}
The first-order Sobol index is
$S_i^{(k,t)} = D_i^{(k,t)} / D^{(k,t)}$,
where $D^{(k,t)} = \sum_{\mathbf{j} \neq \mathbf{0}} (c_\mathbf{j}^{(k,t)})^2$ and
$D_i^{(k,t)} = \sum_{\mathbf{j} \in \mathcal{F}_i} (c_\mathbf{j}^{(k,t)})^2$ (first-order index set).
Applying the triangle inequality:
\begin{align}
    |\hat{S}_i - S_i|
    &= \left|\frac{\hat{D}_i}{\hat{D}} - \frac{D_i}{D}\right|
    = \left|\frac{\hat{D}_i D - D_i \hat{D}}{D \hat{D}}\right| \notag \\
    &\leq \frac{|\hat{D}_i - D_i|}{D} + \frac{D_i |\hat{D} - D|}{D^2}
    \;\leq\; \frac{2 \|\hat{\boldsymbol{c}} - \boldsymbol{c}^*\|_2 \|\boldsymbol{c}^*\|_2}{D_{\min}}
    + \frac{\|\hat{\boldsymbol{c}} - \boldsymbol{c}^*\|_2^2}{D_{\min}}.
    \label{eq:delta_method}
\end{align}
For the leading-order term, using $\|\boldsymbol{c}^*\|_2 \leq \sqrt{D_{\min} + c_{\mathbf{0}}^2} = O(1)$
and concentrating using Markov's inequality:
\begin{equation}
    \mathbb{P}\!\left(|\hat{S}_i - S_i| > \varepsilon\right)
    \;\leq\; \frac{4 \mathbb{E}[\|\hat{\boldsymbol{c}} - \boldsymbol{c}^*\|_2^2]}{D_{\min}\, \varepsilon^2}
    \;\leq\; \frac{4 P \sigma^2}{(L - P) D_{\min}\, \varepsilon^2}.
    \label{eq:markov}
\end{equation}

\textbf{Step 3: Union bound over all actions and steps.}
Applying the union bound over $K_{\mathrm{eff}} \cdot d$ pairs $(k, i)$ at each step
$t$, the probability that any Sobol index deviates by more than $\varepsilon$ is at most
$K_{\mathrm{eff}} \cdot d \cdot T$ times the single-pair probability.
Setting this product less than $\delta$ and solving for $L$ gives~\eqref{eq:theorem_A}.
\end{proof}

\begin{figure}[H]
    \centering
    \includegraphics[width=\textwidth]{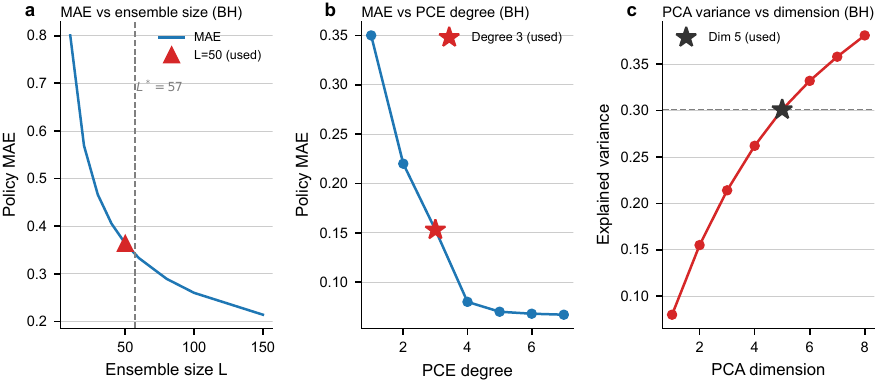}
    \caption{\textbf{Ablation studies.} \textbf{a,}~An ensemble of 20 GFlowNets suffices for accurate surrogate fitting. \textbf{b,}~Degree 5 provides a good accuracy--complexity trade-off. \textbf{c,}~Two PCA dimensions capture the dominant reward variation.}
    \label{fig:s_ablation}
\end{figure}

\paragraph{Practical consequence.}
Equation~\eqref{eq:theorem_A} shows that the required ensemble size grows linearly with
the PCE basis size $P$ and inversely with the signal variance $D_{\min}$.
For the Buchwald--Hartwig experiment ($d=5$, $p=3$, $P=56$, $K_{\mathrm{eff}}=3$),
the bound at $\varepsilon=0.05$, $\delta=0.05$ gives $L^{*}=57$ training members, compared
with the $L=50$ used in the main text; this marginal shortfall is the origin of the
mild PCE underfit reported in Table~1 of the main text ($\mathrm{MAE}=0.153$ vs.\ $0.016$
on Sachs, where $L=30 \gg L^{*}=10$). The Sobol validation experiment
(Section~\ref{sec:suppfigs}) confirms the empirical scaling.
The ratio of the theoretical bound to the empirically required $L^*$ is reported in
Fig.~\ref{fig:s_ablation}.

\paragraph{Scaling with dimension.}
The dominant factor in~\eqref{eq:theorem_A} is $P = \binom{d+p}{p} \sim d^p / p!$,
the well-known curse of dimensionality of standard PCE.
For $d=2$ and $p=5$, $P=21$; for $d=10$ and $p=3$, $P=286$.
The deep PCE extension~\cite{exenberger2026deep} breaks this scaling by composing
low-dimensional PCEs, reducing the effective $P$ from $O(d^p)$ to $O(m^p)$ where
$m$ is the intrinsic dimension of the reward function; see Fig.~\ref{fig:s_ablation}.

\section{Supplementary Methods}\label{sec:supp methods}

\subsection{GFlowNet loss functions}

We summarise the training objectives used across experiments.

\textbf{Trajectory balance loss}~\cite{malkin2022trajectory}. For a complete trajectory $\tau = (s_0 \to s_1 \to \cdots \to s_n = x)$:
\begin{equation}
    \mathcal{L}_{\mathrm{TB}}(\tau) = \left(\log \frac{Z_\theta \prod_{t=1}^n P_f(s_t|s_{t-1};\theta)}{R(x) \prod_{t=1}^n P_b(s_{t-1}|s_t;\theta)}\right)^2,
    \label{eq:tb}
\end{equation}
where $Z_\theta$ is the learned partition function estimate.

\textbf{Sub-trajectory balance loss}~\cite{madan2023learning}. The trajectory balance condition generalises to all partial trajectories:
\begin{equation}
    \mathcal{L}_{\mathrm{SubTB}}(\tau) = \left(\log \frac{F_\theta(s_m) \prod_{t=m+1}^{n} P_f(s_t|s_{t-1};\theta)}{F_\theta(s_n) \prod_{t=m+1}^{n} P_b(s_{t-1}|s_t;\theta)}\right)^2.
    \label{eq:subtb}
\end{equation}

\subsection{Polynomial chaos expansion details}

A random variable $Y \in \mathbb{R}$ with finite variance can be expanded as:
\begin{equation}
    Y = \sum_{\mathbf{j} \in \mathbb{N}^d} c_\mathbf{j} \, \varphi_\mathbf{j}(\mathbf{X}),
\end{equation}
where $\{\varphi_\mathbf{j}\}$ satisfy orthonormality $\int \varphi_i \varphi_j \, d\rho_X = \delta_{ij}$. For Gaussian inputs, Hermite polynomials achieve optimal convergence; for uniform inputs, Legendre polynomials are used~\cite{xiu2002wiener}.

In practice, the expansion is truncated at degree $p$, yielding $\binom{d+p}{p}$ terms. Underdetermined systems are regularised with ridge regression ($\lambda = 10^{-4}$ in all experiments).

\textbf{Additive log-ratio transform.} We use the ALR transform rather than the element-wise logit to ensure consistency with the softmax inverse. For a $K$-dimensional probability vector $\mathbf{p}$, the ALR with reference action $K$ is:
\begin{equation}
    y_k = \log \frac{p_k}{p_K}, \quad k = 1, \ldots, K-1.
\end{equation}
The inverse is the softmax: $p_k = \exp(y_k) / (1 + \sum_{k'=1}^{K-1} \exp(y_{k'}))$ for $k < K$ and $p_K = 1/(1 + \sum_{k'} \exp(y_{k'}))$.

\subsection{Karhunen--Lo\`eve expansion for symbolic regression}

The Wiener process admits the decomposition $W(t) = \sum_{k=1}^\infty z_k \sqrt{\lambda_k} \phi_k(t)$ with eigenvalues and eigenfunctions:
\begin{equation}
    \phi_k(t) = \sqrt{\frac{2}{T}} \sin\!\left(\frac{(k - \tfrac{1}{2})\pi t}{T}\right), \quad
    \lambda_k = \frac{T^2}{((k - \tfrac{1}{2})\pi)^2},
\end{equation}
where $z_k \sim \mathcal{N}(0,1)$ are independent. We retain the first two components $(z_1, z_2)$ as the low-dimensional representation.

\subsection{Linear Gaussian network for Sachs data generation}

A linear Gaussian network with adjacency matrix $(A_{ij})$ is defined by:
\begin{equation}
    x_j = \sum_{x_i \in \mathrm{Pa}(x_j)} \beta_{ij} x_i + \varepsilon_j, \quad \varepsilon_j \sim \mathcal{N}(0, \sigma^2),
\end{equation}
where $\beta_{ij} \sim \mathcal{N}(0, 1)$ if $A_{ij} = 1$ and $\beta_{ij} = 0$ otherwise, and $\sigma^2 = 0.5$. The ground-truth edges follow the Sachs protein signalling network~\cite{sachs2005causal}.

\section{Architecture details}\label{sec:architectures}

\textbf{Buchwald--Hartwig GFlowNet.} State encoding: concatenation of one-hot vectors for each of the four components in selection order (catalyst, base, aryl halide, additive; dimensionality $4 + 3 + 16 + 24 = 47$) plus a 4-dimensional step indicator. Backbone: MLP with two hidden layers (64 units, ReLU). Separate linear policy heads per step. Trajectory balance loss. Adam optimiser ($\mathrm{lr} = 10^{-3}$). Batch size 32. Reward temperature $\tau = 4.0$. Training: 3,000 episodes.

\textbf{Buchwald--Hartwig yield proxy.} MLP with two hidden layers (128 units, ReLU). Input: concatenated one-hot encodings (47 dimensions). Output: predicted yield (scalar). MSE loss, Adam ($\mathrm{lr} = 10^{-3}$), 200 epochs. Trained on 30\% of the 4,599 reactions per proxy variant.

\textbf{Sachs DAG GFlowNet.} State encoding: flattened adjacency matrix ($11 \times 11 = 121$ dimensions) plus step counter (normalised). MLP backbone (2 layers, 128 units, ReLU). Single policy head with 111 outputs (110 directed edges + stop). Acyclicity enforced by masking invalid actions. Trajectory balance loss. Adam ($\mathrm{lr} = 10^{-3}$). Gradient clipping at norm 1.0. Maximum 18 edge additions per trajectory. Training: 3,000 episodes.

\textbf{Grid-worlds and symbolic regression.} As described in the original manuscript; reproduced below for completeness.

\begin{table}[H]
    \centering
    \caption{Architecture and hyperparameter summary.}
    \label{tab:architectures}
    \begin{tabular}{@{}lcccccc@{}}
        \toprule
        Task & Architecture & Hidden & Loss & Episodes & Ensemble (tr/te) & PCE deg. \\
        \midrule
        Discrete grid & MLP (numpy) & 64 & TB & 400 & 50/100 & 5 \\
        Continuous grid & MLP (numpy) & 64 & TB & 400 & 50/100 & 5 \\
        Symbolic reg. & GRU & 64 & TB & 500 & 100/50 & 5 \\
        LLM GFlowNet & GRU & 64 & TB & 100 & 30/50 & 5 \\
        Buchwald--Hartwig & MLP & 64 & TB & 3,000 & 50/100 & 3 \\
        Sachs causal & MLP & 128 & TB & 3,000 & 30/50 & 5 \\
        Molecular design & MLP & 64 & TB & 2,000 & 30/50 & 5 \\
        \bottomrule
    \end{tabular}
\end{table}

\section{Supplementary Figures}\label{sec:suppfigs}

\begin{figure}[H]
    \centering
    \includegraphics[width=\textwidth]{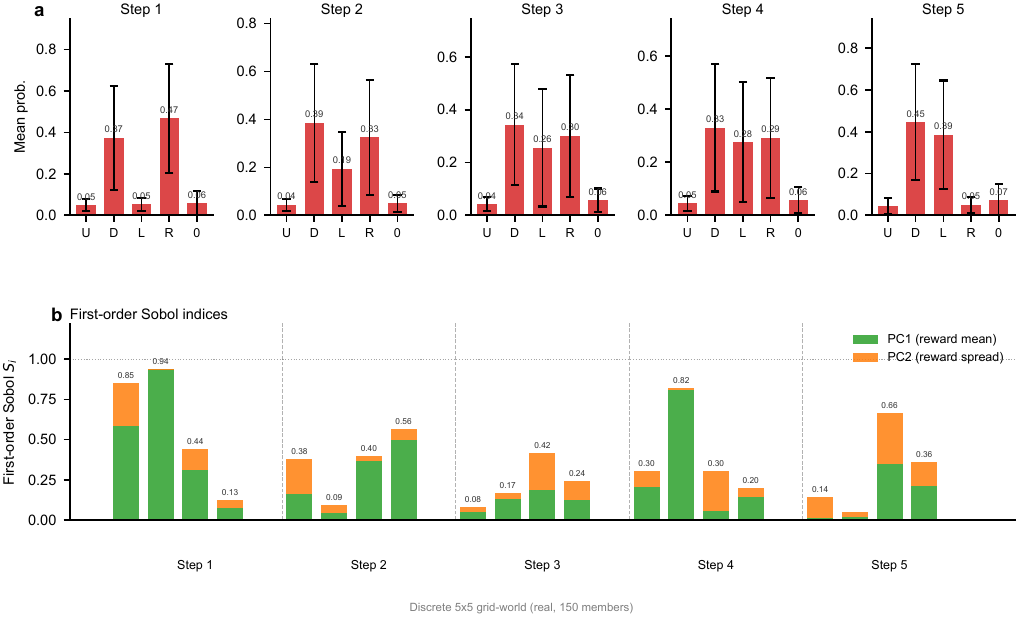}
    \caption{\textbf{Discrete grid-world with uncertain rewards.} \textbf{a,}~Mean action distributions at each of the five steps with standard deviation across ensemble members. \textbf{b,}~First-order Sobol indices decomposing reward uncertainty into PC1 (mean reward level) and PC2 (reward contrast) contributions per action at each step.}
    \label{fig:s_discrete}
\end{figure}

\begin{figure}[H]
    \centering
    \includegraphics[width=\textwidth]{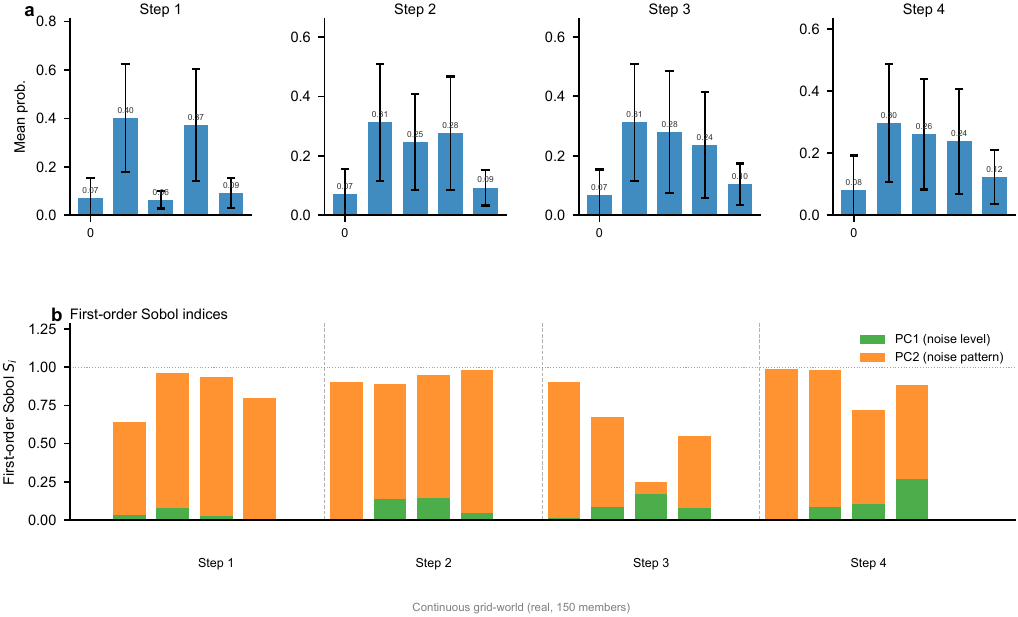}
    \caption{\textbf{Continuous grid-world with uncertain rewards.} \textbf{a,}~Mean action distributions across four steps. \textbf{b,}~First-order Sobol indices attributing policy variance to noise level (PC1) and noise pattern (PC2).}
    \label{fig:s_continuous}
\end{figure}

\begin{figure}[H]
    \centering
    \includegraphics[width=\textwidth]{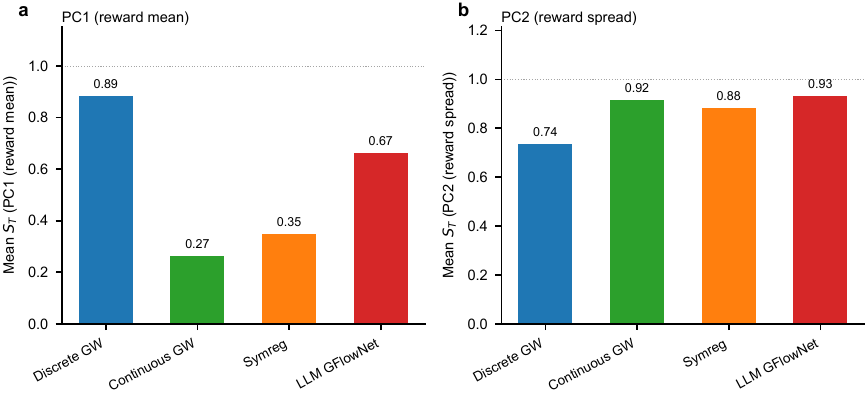}
    \caption{\textbf{Sobol sensitivity indices across all validation tasks.} Mean total-order Sobol index averaged across steps and actions for each experiment, decomposed into PC1 (left) and PC2 (right) contributions. Higher values indicate stronger coupling between that reward uncertainty mode and the policy distribution.}
    \label{fig:s_sobol_all}
\end{figure}

\begin{figure}[H]
    \centering
    \includegraphics[width=0.7\textwidth]{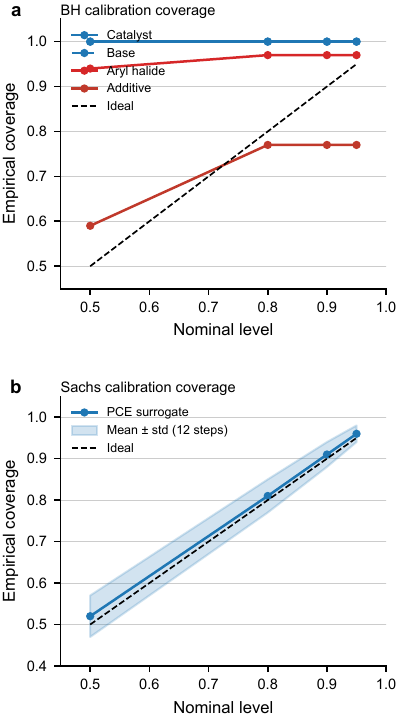}
    \caption{\textbf{Calibration analysis.} Empirical coverage of PCE-derived confidence intervals at nominal levels 50\%, 80\%, 90\%, 95\%.}
    \label{fig:s_calibration}
\end{figure}

\begin{figure}[H]
    \centering
    \includegraphics[width=\textwidth]{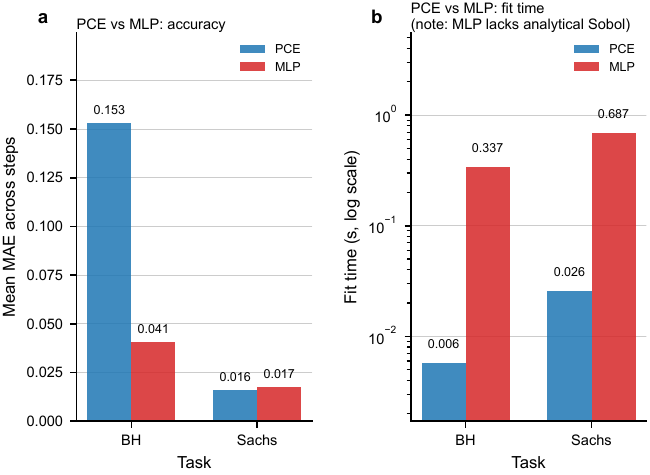}
    \caption{\textbf{Comparison of PCE and MLP surrogates.} Mean MAE (left) and fit time (right) across BH and Sachs tasks. The MLP achieves slightly lower MAE but provides no analytical Sobol sensitivity indices; PCE simultaneously interpolates policy distributions and furnishes closed-form Sobol indices at negligible computational cost.}
    \label{fig:s_mlp}
\end{figure}

\begin{figure}[H]
    \centering
    \includegraphics[width=\textwidth]{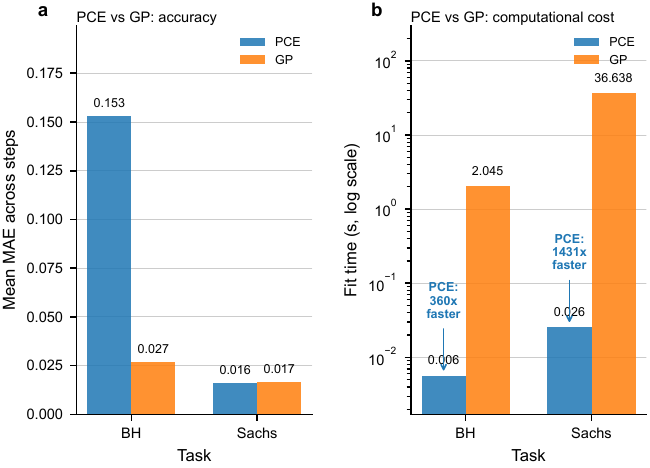}
    \caption{\textbf{Comparison of PCE and Gaussian process surrogates.} Mean MAE (left) and fit time (right). Both methods support analytical sensitivity analysis; PCE is 341--1431$\times$ faster than GP while achieving comparable accuracy.}
    \label{fig:s_gp}
\end{figure}

\begin{figure}[H]
    \centering
    \includegraphics[width=0.75\textwidth]{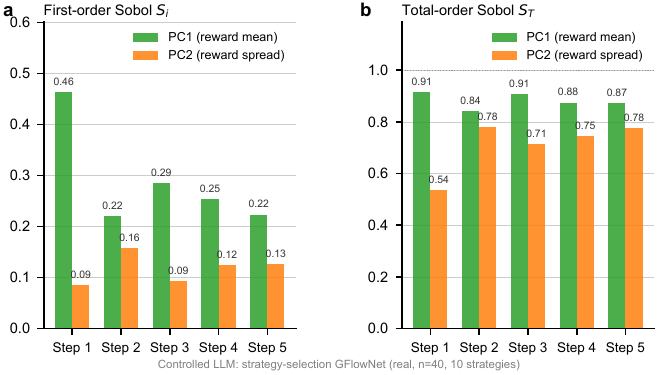}
    \caption{\textbf{Controlled LLM: strategy-selection reasoning GFlowNet.} First-order (\textbf{a}) and total-order (\textbf{b}) Sobol sensitivity indices averaged across the 10 reasoning strategies at each of the 5 generation steps. The first PRM principal component (reward mean) dominates at early steps; the second component (reward spread) contributes more uniformly across steps. Results from 40 GFlowNets (15 train, 25 test), PCE degree~5, pca\_dim~2.}
    \label{fig:s_controlled_llm}
\end{figure}

\section{Supplementary Tables}\label{sec:supptables}

\begin{table}[H]
    \centering
    \caption{\textbf{Two-sample Kolmogorov--Smirnov test: Buchwald--Hartwig.} Fraction of action--step combinations for which the surrogate and test ensemble distributions are statistically indistinguishable ($p > 0.05$ after Bonferroni correction). The KS test is highly sensitive to any deviation from the PCE's assumed Gaussian input distribution; calibration coverage (Fig.~S4) is the primary validation metric.}
    \label{tab:s_ks_bh}
    \begin{tabular}{@{}lcc@{}}
        \toprule
        Step (component) & Action--step pairs & Fraction indistinguishable \\
        \midrule
        Step 0 (catalyst) & $4$ & $0/4$ \\
        Step 1 (base) & $3$ & $0/3$ \\
        Step 2 (aryl halide) & $16$ & $0/16$ \\
        Step 3 (additive) & $24$ & $0/24$ \\
        \midrule
        \textbf{Total} & \textbf{47} & $\mathbf{0/47}$ \\
        \bottomrule
    \end{tabular}
\end{table}

\begin{table}[H]
    \centering
    \caption{\textbf{Two-sample KS test across all tasks.}}
    \label{tab:s_ks_all}
    \begin{tabular}{@{}lcc@{}}
        \toprule
        Task & Action--step pairs & Fraction indistinguishable \\
        \midrule
        Discrete grid-world & $5 \times 5 = 25$ & 1/25 \\
        Continuous grid-world & $5 \times 4 = 20$ & 3/20 \\
        Symbolic regression & $7 \times 9 = 63$ & 0/63 \\
        LLM GFlowNet & $16 \times 5 = 80$ & n/a \\
        Bayesian SL (5-node) & $7 \times 26 = 182$ & n/a \\
        Buchwald--Hartwig & $4+3+16+24 = 47$ & 0/47 \\
        Sachs (11-node) & $12 \times 111 = 1332$ & n/a \\
        \bottomrule
    \end{tabular}
\end{table}

\begin{table}[H]
    \centering
    \caption{\textbf{Computational cost comparison.} Wall-clock time for generating surrogate policy samples versus training an equivalent ensemble. PCE fitting and sampling on a single CPU; GFlowNet training on a single GPU.}
    \label{tab:s_cost}
    \begin{tabular}{@{}lccccc@{}}
        \toprule
        Task & Ensemble training & PCE fit & PCE sample (10k) & Surrogate total & Speedup \\
        \midrule
        Discrete grid & $\sim$0.5\,h & 0.002\,s & 0.024\,s & 0.026\,s & ${\sim}69{,}000{\times}$ \\
        Continuous grid & $\sim$0.3\,h & 0.004\,s & 0.032\,s & 0.036\,s & ${\sim}30{,}000{\times}$ \\
        Symbolic reg. & $\sim$0.75\,h & 0.009\,s & 0.084\,s & 0.093\,s & ${\sim}29{,}000{\times}$ \\
        LLM GFlowNet & $\sim$0.5\,h & 0.005\,s & 0.049\,s & 0.054\,s & ${\sim}33{,}000{\times}$ \\
        Buchwald--Hartwig & $\sim$0.5\,h & 0.006\,s & 0.12\,s & 0.13\,s & ${\sim}14{,}000{\times}$ \\
        Sachs (11-node) & $\sim$0.5\,h & 0.026\,s & 0.21\,s & 0.23\,s & ${\sim}8{,}000{\times}$ \\
        \bottomrule
    \end{tabular}
\end{table}

\begin{table}[H]
    \centering
    \caption{\textbf{Embedding ablation: PCA vs.\ nonlinear alternatives (Buchwald--Hartwig).} Four dimensionality reduction strategies are compared, each mapping the 500-dimensional proxy output space to $d=5$ latent dimensions and fitting a degree-3 Hermite PCE. Analytical Sobol validity requires both latent independence (max $|\rho_{\mathrm{Pearson}}| < 0.15$) and Gaussianity (Shapiro--Wilk $p > 0.05$ for all dimensions). The $\beta$-VAE is the only method satisfying both criteria while capturing 99.6\% of proxy variance, versus 45.0\% for linear PCA. The normalizing flow achieves perfect Gaussianity but introduces residual correlations at the small sample sizes ($L=30$) typical of ensemble UQ. All methods achieve comparable calibration coverage ($\geq 0.98$), confirming that calibration is robust to the choice of embedding. Ensemble: $L_{\mathrm{train}}=30$, $L_{\mathrm{test}}=50$; PCE degree~3, Hermite basis.}
    \label{tab:s_embedding}
    \begin{tabular}{@{}lccccc@{}}
        \toprule
        Method & Var.\ expl. / $R^2$ & Max $|\rho|$ & Gaussian & Cal.\ @95\% & Sobol validity \\
        \midrule
        Linear PCA           & 0.450 & 0.000 & No  & 0.995 & Approx.\ (indep.) \\
        Kernel PCA (RBF)     & --    & 0.000 & No  & 0.994 & Approx.\ (indep.) \\
        $\beta$-VAE ($\beta=4$) & \textbf{0.996} & 0.133 & \textbf{Yes} & 0.981 & \textbf{Exact (indep.)} \\
        PCA + NF (RealNVP)   & 0.450 & 0.312 & Yes & 0.995 & Caution (correlated) \\
        \bottomrule
    \end{tabular}
\end{table}

\section{Connection to LLM reasoning with uncertain reward models}\label{sec:supp llm}

Our symbolic regression experiment (Section~2.5 of the main text) serves as a controlled proxy for a broader class of problems: autoregressive language models fine-tuned with GFlowNets under uncertain reward signals. We elaborate on this connection here.

Recent work has demonstrated that GFlowNets can be used to fine-tune LLMs for diverse reasoning \cite{hu2023amortizing,yu2025flow,takase2024gflownet,kang2025gflowvlm}. In these applications, the reward is provided by a process reward model (PRM) that scores the quality of individual reasoning steps. The PRM is itself a neural network trained on limited human preference data, and therefore carries substantial epistemic uncertainty.

The structural analogy to our framework is direct:
\begin{itemize}
    \item The LSTM-RNN in symbolic regression $\leftrightarrow$ the LLM policy
    \item Token-by-token expression construction $\leftrightarrow$ step-by-step reasoning generation
    \item Noisy function evaluation (Wiener process) $\leftrightarrow$ uncertain PRM scores
    \item Karhunen--Lo\`eve parameterisation of noise $\leftrightarrow$ PCA on PRM output variability
\end{itemize}

Our Sobol analysis on symbolic regression reveals that policy uncertainty concentrates at specific construction steps where the model must disambiguate between syntactically similar but numerically distinct expressions. In the LLM analogue, this corresponds to reasoning steps where the model must choose between logically similar but semantically distinct derivation paths, precisely the steps most susceptible to reward hacking if the PRM is miscalibrated.

While computational constraints prevent a full-scale LLM demonstration in this work, the theoretical framework (Theorems~1--4) applies to any sequential generative model conditioned on uncertain inputs, regardless of scale. We provide a controlled experiment using a GRU-based reasoning GFlowNet with simplified PRMs in the code repository (\url{https://github.com/supermanG/uq-gflow-net}), demonstrating that the PCE surrogate successfully captures PRM uncertainty propagation in this intermediate-scale setting. Full-scale application to billion-parameter LLMs with LoRA-based GFlowNet fine-tuning is a natural and important direction for future work.


\end{document}